\pdfoutput=1

\documentclass[11pt]{article}

\usepackage{acl}
\usepackage{enumitem}

\usepackage{times}
\usepackage{latexsym}

\usepackage[T1]{fontenc}

\usepackage[utf8]{inputenc}

\usepackage{microtype}

\usepackage{inconsolata}

\usepackage{amsmath}
\usepackage[normalem]{ulem}
\usepackage{multirow}
\usepackage{adjustbox}
\usepackage{comment}
\usepackage{bbding}
\usepackage{pifont}
\usepackage{wasysym}
\usepackage{amssymb}
\usepackage{booktabs}
\usepackage{dirtytalk}
\usepackage{graphicx}
\usepackage{color, colortbl}
\usepackage{xcolor, soul}
\usepackage{arydshln}
\usepackage{caption}
\usepackage{subcaption}
\usepackage{supertabular}

\usepackage[textsize=scriptsize]{todonotes}

\definecolor{someSpanText}{HTML}{000000}
\definecolor{colorClassName}{HTML}{F4CCCC}
\definecolor{colorLexicon}{HTML}{D9EAD3}
\definecolor{colorInputSentence}{HTML}{FFF2CC}
\definecolor{randClassForMethod2}{HTML}{ff644e}
\definecolor{randClassForMethod}{HTML}{CC0000}
\definecolor{subsetLexForMethod}{HTML}{38761D}
\definecolor{inputSentForMethod}{HTML}{FF9900}
\definecolor{ourmethodWithInstruction}{HTML}{ff9671}
\definecolor{ourmethodWithoutInstruction}{HTML}{b4b3ff}
\definecolor{nullprompt}{HTML}{dcd5ff}
\definecolor{Rnum}{HTML}{b4b3ff}
\definecolor{RwordMinus}{HTML}{676bea}
\definecolor{Rword}{HTML}{3543bc}
\definecolor{Iprompt}{HTML}{ffd1dc}
\definecolor{IRnum}{HTML}{ffaead}
\definecolor{IRwordMinus}{HTML}{ff9671}
\definecolor{IRword}{HTML}{ff6d01}
\definecolor{lightGrey}{HTML}{E6E6E6}
\definecolor{darkGrey}{HTML}{A7A7A7}

\DeclareRobustCommand{\withInstruct}[1]{\sethlcolor{ourmethodWithInstruction}\hl{#1}}
\DeclareRobustCommand{\withoutInstruct}[1]{\sethlcolor{ourmethodWithoutInstruction}\hl{#1}}
\DeclareRobustCommand{\lightGrey}[1]{\sethlcolor{lightGrey}\hl{#1}}
\DeclareRobustCommand{\darkGrey}[1]{\sethlcolor{darkGrey}\hl{#1}}

\newcommand\ourMethodWithKtuning{+Lex +K} 
\newcommand\sOne{dict}
\newcommand\sTwo{nlp+chat}
\newcommand\sThree{class}
\newcommand\sFour{human}
\newcommand\styleSplitAa{style$_{\text{src1}}$}
\newcommand\styleSplitA{style$_{\text{src2}}$}
\newcommand\styleSplitB{style$_{\text{src3}}$}
\newcommand\styleSplitC{style$_{\text{src4}}$}
\newcommand\noLexMethod{Standard}
\newcommand\LexPTmethod{+ Lex}

\makeatletter
\def\adl@drawiv#1#2#3{%
        \hskip.5\tabcolsep
        \xleaders#3{#2.5\@tempdimb #1{1}#2.5\@tempdimb}%
                #2\z@ plus1fil minus1fil\relax
        \hskip.5\tabcolsep}
\newcommand{\cdashlinelr}[1]{%
  \noalign{\vskip\aboverulesep
           \global\let\@dashdrawstore\adl@draw
           \global\let\adl@draw\adl@drawiv}
  \cdashline{#1}
  \noalign{\global\let\adl@draw\@dashdrawstore
           \vskip\belowrulesep}}
\makeatother

\title{Meta-Tuning LLMs to Leverage Lexical Knowledge for Generalizable Language Style Understanding}

\newcommand{\AnD}{\hskip 2em plus 1fil minus 0.5em}
 \author{Ruohao Guo \AnD Wei Xu \AnD Alan Ritter \\
 Georgia Institute of Technology \\
 \texttt{rguo48@gatech.edu; \{wei.xu, alan.ritter\}@cc.gatech.edu}
 }

\begin{document}
\maketitle
\begin{abstract}
Language style is often used by writers to convey their intentions, identities, and mastery of language. In this paper, we show that current large language models struggle to capture some language styles without fine-tuning. To address this challenge, we investigate whether LLMs can be meta-trained based on representative lexicons to recognize new styles they have not been fine-tuned on. Experiments on 13 established style classification tasks, as well as 63 novel tasks generated using LLMs, demonstrate that meta-training with style lexicons consistently improves zero-shot transfer across styles. We release the code and data at \url{https://github.com/octaviaguo/Style-LLM}.

\vspace{2pt}
\end{abstract}

\section{Introduction} 
\begin{figure*}[t]
\centering
\vskip -0.1in
    \includegraphics[width=.95\textwidth]{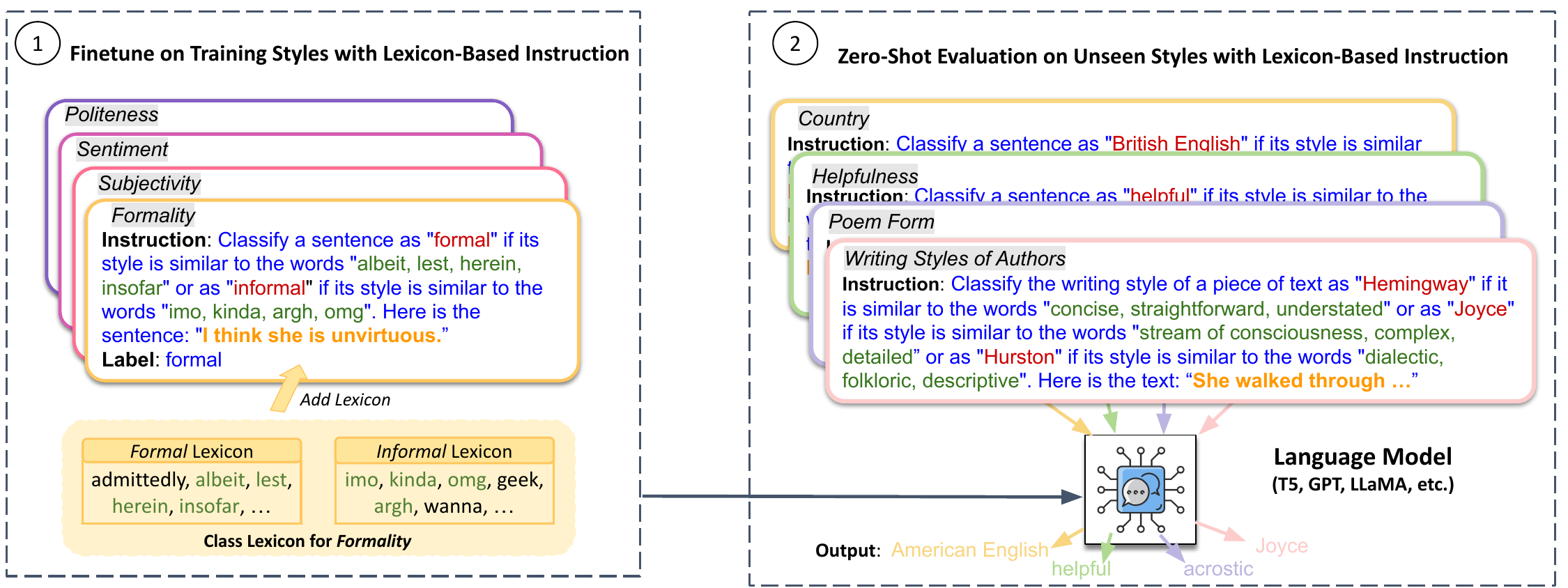}
    \caption{
    Overview of using lexicon-based instructions for cross-style zero-shot classification. It consists of two steps: (1) instruction tuning the model on training styles; (2) evaluating the learned model on unseen target styles zero-shot. A lexicon-based instruction is composed of \textcolor{blue}{instruction}, \textcolor{randClassForMethod}{class names}, \textcolor{subsetLexForMethod}{lexicons} and \textcolor{inputSentForMethod}{an input}.
        }
    \label{fig:lexicon_based_promopt_method}
    \vspace{-.1cm}
\end{figure*}

The style of a text refers to unique ways authors select words and grammar to express their message \cite{hovy1987generating}, providing insights into social interactions and implicit communication. The open-ended and ever-evolving nature of style \cite{xu-2017-shakespeare,kang2021xslue} motivates the need for zero-shot classification, as it is costly to annotate data for every possible style in every language.

\textcolor{black}{Recent large language models (LLMs) and their instruction-tuned variants have achieved notable success in zero-shot learning for diverse tasks using prompting strategies \cite{brown2020language}. Yet, their efficacy in style classification has not been thoroughly investigated. As we will show in this paper (\S \ref{main_experimental_results}), style classification remains a challenge for standard LLM prompting.} On the other hand, before the paradigm in NLP shifted to pre-trained language models, lexicons of words that are stylistically expressive were commonly used as important lexical knowledge \cite{verma2019lexical} in rule-based \citep{wilson-etal-2005-recognizing,taboada2011lexicon}, feature-based \citep{mohammad2013nrc,eisenstein2017unsupervised}, and deep learning models \citep{teng-etal-2016-context,maddela-xu-2018-word} for style identification. Many lexicons have been developed for varied styles, such as politeness \citep{politeness2013}, happiness \cite{dodds2015human}, emotions \citep{mohammad2010emotions,tausczik2010psychological}, etc. This leads to a natural question: \textit{can we leverage lexicons during instruction fine-tuning of LLMs to improve their understanding of language style?}

In this paper, we examine the effectiveness of fine-tuning LLMs to interpret lexicons that are provided as inputs to elicit latent knowledge \citep{kang2023self} of language styles that were acquired during pre-training. We first compile a benchmark of 13 diverse writing styles with both annotated test sets and style-representative lexicons.  Using this benchmark, we show that \textbf{meta-tuning with lexicons} enables different pre-trained LLMs to generalize better to new styles that have no labeled data. For example, meta-tuning LLaMA-2-7B \cite{touvron2023llama} on seven styles can improve the average F1 score on a separate set of six held-out styles by 12\%, and by 8\% over a general instruction-tuned model, LLaMA-2-Chat.

To further verify the capability of LLMs to generalize to novel styles using lexicons as the only source of supervision, we generated a diverse set of 63 unique writing styles with examples (\S \ref{sec:gpt4-styles}) using an approach inspired by self-instruction \citep{wang-etal-2023-self-instruct}. We demonstrate that using a small lexicon of as few as five words can effectively improve generalization to new styles. We found it helpful to replace class names with random identifiers when meta-training models with lexicons, which prevents models from ignoring the lexicons and simply memorizing source styles' class names. In addition, we show that when combined with \textbf{meta in-context learning} \cite{min-etal-2022-metaicl,flant5}, incorporating lexicons can significantly reduce variance.

\textcolor{black}{In summary, our contributions are: 
(1) we introduce lexicon-based instructions (\S\ref{section:meta_tuning_setup}), a simple yet effective method for zero-shot style classification leveraging lexical knowledge in LLMs; 
(2) we show class randomization (\S\ref{section:meta_tuning_setup}) can improve generalization capability of lexicon-instructed models significantly (\S\ref{lexicon-based-prompts});
(3) we provide a benchmark for zero-shot style classification, featuring 13 established tasks (\S\ref{section:experimental_setup}) and a synthetic dataset of 63 new  tasks (\S\ref{sec:gpt4-styles}), complete with representative lexicons.
}

\begin{table*}[t]
\centering
\small
\vskip -0.1in
\setlength{\tabcolsep}{2pt}
\begin{adjustbox}{width=\linewidth}
\begin{tabular}{lcclll} 
\toprule
\textbf{Style Dataset} & \textbf{$|C|$} & \textbf{B?}             & \textbf{Domain}& \textbf{\#Tra, Val, Test} & \textbf{Lexicon Sources}  \\ 
\midrule
Age$^*$ \cite{kang2021xslue}   & 2 & \ding{55} & caption &  14k, 2k, 2k     & ChatGPT, Dict        \\ 
\rowcolor[HTML]{EFEFEF} Country   \cite{kang2021xslue}   & 2   & \ding{55} & caption   &  33k, 4k, 4k  & ChatGPT, Dict     \\
Formality \cite{formality2018}    & 2      & \checkmark & web      &    209k,10k,5k    & NLP \cite{formalityLexicon2010}, Dict       \\ 
\rowcolor[HTML]{EFEFEF} Hate/Offense \cite{davidson2017automated}          & 3         & \ding{55}  & Twitter &    22k,1k,1k        & NLP \cite{hateoffense-nlp-lexicon}, Dict      \\ 
Humor \cite{shorthumor}           & 2                    & \checkmark        &    web &  40k,2k,2k       & ChatGPT, Dict           \\
\rowcolor[HTML]{EFEFEF} Politeness \cite{politeness2013}       & 2    & \checkmark    & web &   10k,0.5k,0.6k  & NLP \cite{politeness2013}, Dict    \\ 
Politics   \cite{kang2021xslue}         & 3                & \ding{55}   &  caption &  33k, 4k, 4k  & NLP \cite{politics_public_lex}, Dict   \\ 
\rowcolor[HTML]{EFEFEF} Readability  \cite{cefr}  & 2 & \ding{55} & web, Wiki  & 7k,1k,1k  & NLP \cite{maddela-xu-2018-word}, Dict         \\ 
Romance   \cite{kang2021xslue}   & 2    & \checkmark   &   web & 2k,0.1k,0.1k  &   ChatGPT, Dict         \\  
\rowcolor[HTML]{EFEFEF} Sarcasm \cite{sarc}     & 2    & \checkmark &  Reddit  &  11k,3k,3k    & ChatGPT, Dict           \\ 
Sentiment  \cite{socher-etal-2013-recursive}   & 2  & \ding{55}   & web  & 236k,1k,2k  & NLP \cite{SentimentEmotionSurvey2021}, Dict        \\ 
\rowcolor[HTML]{EFEFEF} Shakespeare  \cite{xu2012paraphrasing}       & 2      & \checkmark  & web  &  32k,2k,2k  & NLP \cite{xu2012paraphrasing}, Dict     \\ 
Subjectivity \cite{subjectivity}     & 2    & \checkmark        &  web  &    6k,1k,2k  &   NLP \cite{wilson-etal-2005-recognizing}, Dict          \\ 
\bottomrule
\end{tabular}
\end{adjustbox}
\caption{Statistics of datasets and lexicons. ``$|C|$" denotes the number of classes in each style dataset. ``B?" indicates whether or not the class distribution is balanced. ``\#Tra, Val, Test" lists the number of examples in train, validation and test sets. 
To better compare across different styles, we mapped the original eight classes (i.e., \underline{\emph{Under12}}, \underline{\emph{12-17}}, \underline{\emph{18-24}}, \emph{25-34}, \emph{35-44}, \uwave{\emph{45-54}}, \uwave{\emph{55-74}}, \uwave{\emph{75YearsOrOlder}}) in \emph{Age} dataset into two new classes (i.e., \underline{\emph{\smash{youthful}}}, \uwave{\emph{mature}}). 
}
\label{tab:dataset_statistics}
\end{table*}

\section{Meta-Tuning for Style Generalization}
We investigate the capabilities of LLMs to interpret language styles using lexical knowledge, and identify text that is representative of the associated styles.  
\textcolor{black}{We compare lexicon-based instructions with other methods in the zero-shot setting, and further explore a few-shot setting.}
To study the effectiveness of meta-tuning with lexicons, in generalizing to various writing styles, we first consider a set of thirteen styles, where high-quality annotated data is available.  Later, in \S \ref{sec:gpt4-styles}, we further demonstrate the ability of lexicon-instructed models to generalize using 63 novel LLM-generated styles.

\subsection{Problem Definition and Approach}
\label{section:meta_tuning_setup}

Given an input text and style with pre-defined classes $C=\left \{c_k \right \}^{|C|}_{k=1}$, we present the language model with lexicon-based instructions, by instantiating lexicons (i.e., a list of words or phrases that are representative of each class $c_k$) in a pre-defined instruction template (see templates in Table \ref{tab:prompts_examples}).
A language model is expected to predict one of the classes $\hat{c}\in C$, given the lexicon-based instruction.
These style-lexicons, are the only source of target-style supervision provided to the LLM, enabling it to make stylistic predictions using parametric knowledge that was acquired during pre-training \cite{2020t5, brown2020language}, and elicited using lexicon-based instructions.

\paragraph{Meta-Tuning on Source Styles.}
\label{method:instruction_ft}
Meta-tuning is a simple, but effective approach that directly optimizes the zero-shot learning objective through fine-tuning on a collection of datasets \citep{zhong2021adapting}. \footnote{In this work, we use the term {\em meta-tuning} to mean {\em fine-tuning} on diverse datasets and tasks. {\em Instruction tuning} further narrows the scope to datasets that include instructions. }
In order to guide models to draw upon latent lexical knowledge for zero-shot style classification, we meta-tune LLMs to learn to understand style-lexicon relations.
During preliminary experiments, we found that it is important to make use of \textit{class randomization} (\S \ref{variant_lps}) during meta-tuning, e.g. using randomly selected identifiers to replace more meaningful style labels (e.g., ``humorous''), in order to prevent models from simply memorizing the (source) styles used for fine-tuning.  Without randomizing labels, memorization prevents the model from effectively generalizing to interpret lexicons for new styles. 
In \S\ref{lexicon-based-prompts}, we conduct an analysis into the impact of randomization, and compare different types of identifier randomization.

\paragraph{Zero-Shot Evaluation on Unseen Target Styles.}
To make predictions, we provide the model with the target-style lexicon and use rank classification \cite{t0}, in which we compute the likelihood of each style label, and then pick the one with highest likelihood as the final prediction.

\subsection{A Benchmark for Style Generalization}
\label{section:experimental_setup}

\paragraph{Style Datasets.} We include thirteen language styles that have sentence-level annotated datasets available, covering a wide range of domains, as summarized in Table \ref{tab:dataset_statistics}. These come from a variety of sources, including the XSLUE benchmark \cite{kang2021xslue}, Subjectivity \cite{subjectivity}, Shakespeare \cite{xu2012paraphrasing} and Readability \cite{cefr} - more details are available in Appendix \ref{dataset_selection_from_xslue}.
In the cross-style zero-shot setting, a model is fine-tuned on a set of training styles, then evaluated on a separate set of held-out styles with no overlap. For each training style, its training set is used for fine-tuning, and the validation set is used for model selection \cite{chen2021model}. 
We ensure evaluation style datasets do not share any examples with training styles. Given space limitations, we present results for one split, which includes Sentiment, Formality, Politeness, Hate/Offense, Readability, Politics, and Subjectivity in the training split, while the remaining six styles are included in evaluation split. Experiments on more style splits are shown in Appendix \ref{appendix:style_splits}.

\paragraph{Lexicon Collection.} \label{lexicon_collection} We use stylistic lexicons that have been created by other NLP researchers where possible (listed as ``NLP'' in Table \ref{tab:dataset_statistics}). These lexicons were either manually annotated \cite{maddela-xu-2018-word} or automatically induced using corpus-based approaches  \cite{politeness2013}. 
For styles where such lexicons are not readily available, we explore three methods to create lexicons: (i) prompting ChatGPT to generate words for each class of a style, e.g., the words for the ``humorous'' class are ``funny, laugh-out-loud, silly''; (ii) extracting the definition of each class from Google Dictionary,\footnote{An online service licensed from Oxford University Press: \url{https://www.google.com/search?q=Dictionary}} e.g., ``being comical, amusing, witty" for the ``humorous'' class; (iii) asking a native English speaker to write a list of words for each style. Creation details and more lexicon examples are provided in Appendix \ref{appendix:lexicon_all_details}.

\subsection{Experimental Settings}
\label{subset:methods}
To assess the effectiveness and generality of lexicon-based instructions, we compare our approach with other prompting methods in two learning settings.

\subsubsection{Zero-Shot}
\label{exp_setup:zero-shot}
A model is prompted to predict the evaluation styles without any labeled data. In this setting, we evaluate our Style-* models that are instruction-tuned on the training styles (introduced in \S \ref{section:meta_tuning_setup}). We also experiment with models fine-tuned on general instruction tuning data, including Flan-T5 and LLaMA-2-Chat. 
For each model, we compare the \textbf{Standard} instructions and lexicon-based instructions (i.e., \textbf{\LexPTmethod})  without demonstrations (i.e., example sentences for a evaluation style). 
Both methods utilize the same instruction template described in Appendix \ref{appendix:main_exp_template}, except that class names instead of lexicons are used in standard instructions. To construct a lexicon-based instruction, for each class (e.g., ``polite'' or ``impolite'') of the style (e.g., politeness), we randomly select $m$ words from the corresponding lexicon, then incorporate them into the instruction. We use $m=5$ in the main paper and perform an analysis on varied values of $m$ in Appendix \ref{appendix:lexicon_size}.

\subsubsection{Few-Shot}
We also investigate how different prompting methods perform in the few-shot setting, where a few training examples of the evaluation styles are available. These experiments are not necessarily intended to improve upon the state-of-the-art on this benchmark, but rather to compare the impacts of using in-context examples versus lexicons in enhancing few-shot generalization capabilities.

\paragraph{MetaICL~\cite{min-etal-2022-metaicl,flant5}.} \hspace{-0.06in}\textcolor{black}{We adapt MetaICL, a method that meta-tunes on a collection of tasks with an in-context learning objective, to align models to better learn from the in-context examples through meta-tuning on a set of source styles.} 
During each iteration of fine-tuning, one source style is sampled, and $K$ labeled examples are randomly selected from the train set of that style.  
Each prompt consists of $K$ demonstrations followed by an input sentence for the model to predict the class.
At inference time, 
the prompt is built similarly to the fine-tuning stage, except that the $K$ demonstrations are sampled from the train set of target styles instead of source styles. 
Recently, \citet{2022RethinkingL} have shown that ground-truth labels are not always needed in MetaICL. We re-examine this finding in the context of style classification, experimenting with both random and gold labels in the demonstrations.
We follow \citet{2022RethinkingL} to set $K=4$ and $K=16$. 

\paragraph{MetaICL+Lex.} For a more comprehensive comparison between the two sources of supervision (i.e., demonstrations vs. lexicons), we also modify MetaICL to incorporate lexicons. Specifically, we concatenate the name of each class with its corresponding lexicon words, and prepend this information to each labeled example to form a modified demonstration. 
Each prompt contains $K$ modified demonstrations followed by an input sentence.

\begin{table*}[t]
    \centering
    \vskip -0.08in
    \begin{adjustbox}{width=\linewidth}
    \begin{tabular}{lclccccccc}
    \toprule
    \multirow{1}{*}{\textbf{Model}} &\textbf{Meta-Tuned?} & \multirow{1}{*}{\textbf{\begin{tabular}[c]{@{}l@{}}\textcolor{black}{Instruction}\end{tabular}}} &
            \text{Shakespeare} & \text{Romance} & \text{Humor} & \text{Country} & \text{Sarcasm} & \text{Age} & \textbf{Avg.} \\ \midrule
    \multirow{2}{*}{Flan-T5$_\text{base}$}   & \ding{55}  & \noLexMethod                          & 33.36                & 33.33            & 33.33          & 43.15            & 33.33            & 33.92        & 35.07       \\
                                    &\cellcolor[HTML]{EFEFEF}\ding{55}&\cellcolor[HTML]{EFEFEF}{\LexPTmethod}$_{\text{}}$                            & \cellcolor[HTML]{EFEFEF}49.95               & \cellcolor[HTML]{EFEFEF}51.30            & \cellcolor[HTML]{EFEFEF}48.66          & \cellcolor[HTML]{EFEFEF}35.34            & \cellcolor[HTML]{EFEFEF}49.40           & \cellcolor[HTML]{EFEFEF}49.02        & \cellcolor[HTML]{EFEFEF}\textbf{47.28}                   \\
    \multirow{2}{*}{Style-T5$_\text{base}$}  &\checkmark & \noLexMethod                         & 33.31                & 43.57            & 36.43          & 19.86            & 33.37            & 35.75        & 33.72        \\
                                    &\cellcolor[HTML]{EFEFEF}\checkmark & \cellcolor[HTML]{EFEFEF}{\LexPTmethod}$_{\text{}}$                            & \cellcolor[HTML]{EFEFEF}55.10                & \cellcolor[HTML]{EFEFEF}78.98            & \cellcolor[HTML]{EFEFEF}60.56          & \cellcolor[HTML]{EFEFEF}49.09            & \cellcolor[HTML]{EFEFEF}49.25            & \cellcolor[HTML]{EFEFEF}50.80        & \cellcolor[HTML]{EFEFEF}\textbf{57.30}                   \\\midrule
    \multirow{2}{*}{Style-GPT-J}    &\checkmark&\noLexMethod                        & 58.16                & 87.82            & 33.33          & 53.11            & 44.10            & 35.25        & 51.96          \\
                                    &\cellcolor[HTML]{EFEFEF}\checkmark& \cellcolor[HTML]{EFEFEF}{\LexPTmethod}$_{\text{}}$                            & \cellcolor[HTML]{EFEFEF}56.76                & \cellcolor[HTML]{EFEFEF}83.99            & \cellcolor[HTML]{EFEFEF}55.86          & \cellcolor[HTML]{EFEFEF}44.97            & \cellcolor[HTML]{EFEFEF}48.84            & \cellcolor[HTML]{EFEFEF}47.47        & \cellcolor[HTML]{EFEFEF}\textbf{56.32}                              \\\midrule
\multirow{2}{*}{\begin{tabular}[c]{@{}l@{}}LLaMA-$\text{2-Chat}$\\ \hspace{0.2in} (7B)\end{tabular}}&\ding{55}
& \noLexMethod   &   60.20   & 85.72 & 43.84 & 49.19 & 36.02 & 38.91 & 52.31\\
                                    &\cellcolor[HTML]{EFEFEF}\ding{55}& \cellcolor[HTML]{EFEFEF}{\LexPTmethod}$_{\text{}}$                              & \cellcolor[HTML]{EFEFEF}62.59        & \cellcolor[HTML]{EFEFEF}88.95      & \cellcolor[HTML]{EFEFEF}51.01 & \cellcolor[HTML]{EFEFEF}50.88      & \cellcolor[HTML]{EFEFEF}42.88     & \cellcolor[HTML]{EFEFEF}36.54   & \cellcolor[HTML]{EFEFEF}\textbf{55.47}                      \\\cdashlinelr{1-10}
    \multirow{2}{*}{\begin{tabular}[c]{@{}l@{}}LLaMA-$\text{2-Chat}$\\ \hspace{0.2in} (13B)\end{tabular}} &\ding{55} &\noLexMethod                          & 61.99  & 97.00  & 47.42  & 17.96 & 43.26  & 48.16 & 52.63  \\
                                    &\cellcolor[HTML]{EFEFEF}\ding{55}&\cellcolor[HTML]{EFEFEF}{\LexPTmethod}$_{\text{}}$      & \cellcolor[HTML]{EFEFEF}63.49          & \cellcolor[HTML]{EFEFEF}95.00        & \cellcolor[HTML]{EFEFEF}55.15         & \cellcolor[HTML]{EFEFEF}24.41          & \cellcolor[HTML]{EFEFEF}44.66        & \cellcolor[HTML]{EFEFEF}53.88      & \cellcolor[HTML]{EFEFEF}\textbf{56.10}                       \\ \cdashlinelr{1-10}
    \multirow{2}{*}{\begin{tabular}[c]{@{}l@{}}LLaMA-$\text{2}$\\ \hspace{0.2in} (7B)\end{tabular}}& \ding{55}& \noLexMethod & 42.13 & 64.41 & 37.38 & 48.27 & 48.84 &37.13 & 46.36\\
                                    &\cellcolor[HTML]{EFEFEF}\ding{55} &\cellcolor[HTML]{EFEFEF}{\LexPTmethod}$_{\text{}}$      & \cellcolor[HTML]{EFEFEF}50.21     & \cellcolor[HTML]{EFEFEF}77.86    & \cellcolor[HTML]{EFEFEF}45.44       & \cellcolor[HTML]{EFEFEF}49.86         & \cellcolor[HTML]{EFEFEF}47.72      & \cellcolor[HTML]{EFEFEF} 47.63  & \cellcolor[HTML]{EFEFEF}\textbf{53.12}                       \\ \cdashlinelr{1-10}
    \multirow{2}{*}{\begin{tabular}[c]{@{}l@{}}Style-LLaMA\\ \hspace{0.2in} (7B)\end{tabular}} &\checkmark &\noLexMethod        & 40.91                & 41.65            & 48.88          & 48.92            & 49.02           & 49.80        & 46.53        \\
                                    &\cellcolor[HTML]{EFEFEF}\checkmark  &\cellcolor[HTML]{EFEFEF}{\LexPTmethod}$_{\text{}}$      & \cellcolor[HTML]{EFEFEF}59.03              & \cellcolor[HTML]{EFEFEF}88.97         & \cellcolor[HTML]{EFEFEF}57.64          & \cellcolor[HTML]{EFEFEF}51.52           & \cellcolor[HTML]{EFEFEF}50.83          & \cellcolor[HTML]{EFEFEF}50.53       & \cellcolor[HTML]{EFEFEF}\textbf{59.75}                       \\ \bottomrule
    \end{tabular}
    \end{adjustbox}\vskip -0.03in
    \caption{Zero-shot performance (F1) on the unseen evaluation styles. 
    We compare the models fine-tuned on general instruction tuning data (i.e., not meta-tuned) and the ``Style-*'' models that are instruction-tuned on our training styles (i.e., meta-tuned). For each model, we evaluate its zero-shot learning capabilities when the standard and lexicon-based instructions are used, respectively. 
    } 
    \label{tabs:main_exp}
\end{table*}

\begin{table*}[]
\centering
\small
\begin{adjustbox}{width=\linewidth}
\begin{tabular}{lccccccc}
\toprule
                    
\textbf{Baseline Method}& Shakespeare                                          & Romance                                              & Humor                                                & Country                                              & Sarcasm                                              & Age                                                  & \textbf{Avg.}                                        \\ \midrule
        Majority Classifier                        & 33.30                                                & 33.30                                                & 33.30                                                & 49.20                                                & 33.30                                                & 35.30                                                & 36.28                                                                \\
        Lex Frequency                          &59.91$_{83\%}$ &32.89$_{28\%}$ &33.33$_{0.49\%}$ &50.79$_{5.7\%}$ &33.33$_{0.59\%}$ &37.85$_{18\%}$ &41.35\\
        Lex Emb Sim (Word2Vec)                   & \cellcolor[HTML]{FFFDFA}{\color[HTML]{333333} 49.06} & \cellcolor[HTML]{FFFDFA}{\color[HTML]{333333} 33.33} & \cellcolor[HTML]{FFFDFA}{\color[HTML]{333333} 33.54} & \cellcolor[HTML]{FFFDFA}{\color[HTML]{333333} 49.30} & \cellcolor[HTML]{FFFDFA}{\color[HTML]{333333} 33.33} & \cellcolor[HTML]{FFFDFA}{\color[HTML]{333333} 50.84} & \cellcolor[HTML]{FFFDFA}{\color[HTML]{333333} 41.57}                      \\
        Lex Emb Sim (SentenceBert)               & \cellcolor[HTML]{FFFDFA}{\color[HTML]{333333} 52.00} & \cellcolor[HTML]{FFFDFA}{\color[HTML]{333333} 69.81} & \cellcolor[HTML]{FFFDFA}{\color[HTML]{333333} 57.62} & \cellcolor[HTML]{FFFDFA}{\color[HTML]{333333} 31.12} & \cellcolor[HTML]{FFFDFA}{\color[HTML]{333333} 47.91} & \cellcolor[HTML]{FFFDFA}{\color[HTML]{333333} 49.96} & \cellcolor[HTML]{FFFDFA}{\color[HTML]{333333} 51.40}                      \\
        LLaMA-2 (7B) 16-shot ICL       &  72.16$_{\pm6.94}$              &   57.58$_{\pm21.12}$                                                  &   49.21$_{\pm8.27}$                                                &  43.45$_{\pm10.51}$                                            &  35.21$_{\pm3.29}$                                               &  39.03$_{\pm7.39}$                                                   &        49.44                                                         \\ \bottomrule
\end{tabular}
\end{adjustbox}\vskip -0.03in
\caption{Performance (F1) of baselines.  We compare four approaches: (1) The majority classifier, which predicts the majority label in training data. (2) The lexicon frequency baseline, which counts the occurrence of words from an input sentence in each class's lexicon and then predicts the class with the highest count; the subscript on the score reflects the lexicon usage, i.e., the percentage (\%) of evaluation data that contains at least one word from the corresponding lexicons. (3) The lexicon embedding similarity method, which calculates the cosine similarity between the embeddings of lexicon words for each class and an input, predicting the class with the highest similarity. (4) \textcolor{black}{The In-Context Learning method (without meta-tuning), where a set of 16 training examples for each evaluation style are provided within a prompt for evaluation style prediction; examples are sampled with five random seeds.} 
}
\label{tabs:main_exp_baseline}
\end{table*}

\begin{table*}[t]
\vskip-0.1in
\begin{adjustbox}{width=\linewidth}
\centering
\small
\begin{tabular}{llccccccc}
\toprule
         & \textbf{Method}    & Shakespeare & Romance & Humor & Country & Sarcasm & Age   & \textbf{Avg.} \\ \midrule
\multicolumn{1}{l}{\multirow{3}{*}{\begin{tabular}[c]{@{}l@{}} Examples w/ \\ \emph{random} labels\end{tabular}}} & MetaICL$_4$    & 44.37$_{\pm6.99}$       & 56.21$_{\pm26.64}$   & 37.82$_{\pm5.02}$ & 41.84$_{\pm18.46}$   & 35.55$_{\pm2.94}$   & 40.96$_{\pm11.19}$ & 42.79     \\
\multicolumn{1}{l}{}    &  MetaICL$_4$+Lex      & 39.80$_{\pm1.47}$       & 64.58$_{\pm18.72}$   & 38.59$_{\pm4.41}$ & 49.72$_{\pm0.44}$  & 43.77$_{\pm6.52}$   & 35.30$_{\pm0.00}$ & 45.29\\
\multicolumn{1}{l}{}    &MetaICL$_{16}$ &    55.49$_{\pm11.66}$  &66.91$_{\pm20.48}$ & 36.11$_{\pm4.58}$ &7.74$_{\pm4.67}$ &33.33$_{\pm0.00}$   &31.24$_{\pm0.00}$ &38.47         \\
\midrule
\multicolumn{1}{l}{\multirow{3}{*}{\begin{tabular}[c]{@{}l@{}} Examples w/ \\ \emph{gold} labels\end{tabular}}}   & MetaICL$_4$  & 64.30$_{\pm13.01}$       & 53.53$_{\pm27.30}$   & 49.79$_{\pm12.46}$ & 49.29$_{\pm0.01}$   & 34.28$_{\pm1.57}$   & 36.21$_{\pm1.25}$ & 47.90      \\
\multicolumn{1}{l}{}    &MetaICL$_4$+Lex      &43.90$_{\pm8.06}$       &75.80$_{\pm6.52}$   & 42.78$_{\pm3.99}$ &49.42$_{\pm0.36}$  &38.62$_{\pm3.69}$  & 35.30$_{\pm0.00}$ & 47.63      \\
\multicolumn{1}{l}{}    &  MetaICL$_{16}$          & 72.93$_{\pm8.15}$       & 95.79$_{\pm0.84}$   & 52.05$_{\pm8.52}$& 47.90$_{\pm3.07}$ &  33.33$_{\pm0.00}$  &  35.30$_{\pm0.00}$ & 56.22    \\\bottomrule
\end{tabular}
\end{adjustbox}\vskip-0.01in
\caption{Few-shot performance (F1) of GPT-J. The subscript of MetaICL indicates the number ($K$) of demonstrations  in one prompt. 
For each method (MetaICL$_\text{K}$, or MetaICL$_\text{K}$+Lex), we choose a set of $K$ examples with five different random seeds. 
More results on varying values of $K$ are shown in Appendix \ref{appendix:ablation_k}. We also modify lexicon-based instructions for few-shot learning and compare it with other few-shot learning methods in Appendix \ref{appendix:ourMethodWithKtuning}.
}
\label{tabs:few-shot-exp}
\end{table*}    
\begin{figure*}[]
    \centering \vskip-0.08in
    \includegraphics[width=2\columnwidth]{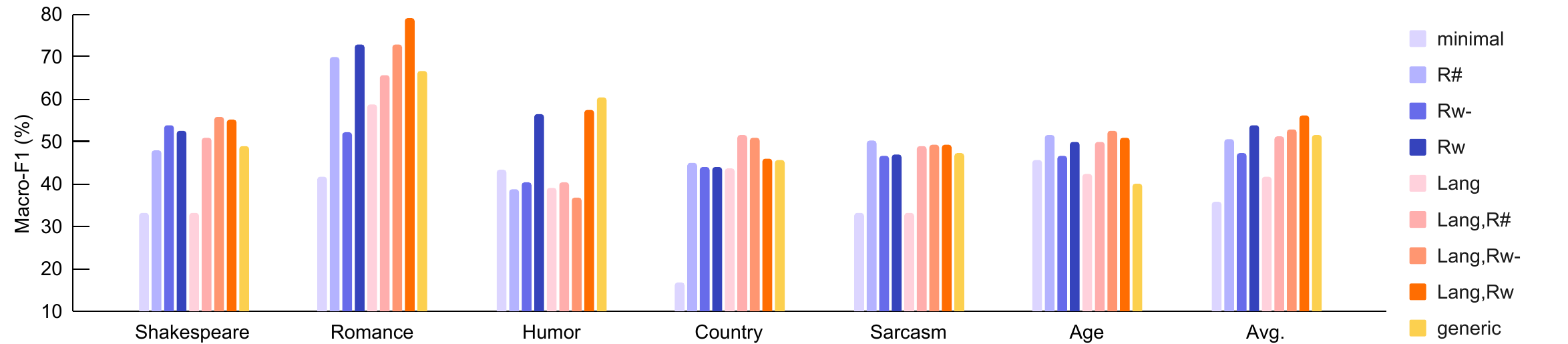}
    \caption{Zero-shot performance when fine-tuning with different lexicon-based instruction variants. Instruction tuning with class \underline{R}andomization shows advantages over those without. Instructions \withInstruct{with} natural language perform generally better than those \withoutInstruct{without}. \textcolor{black}{We also compare fine-tuning with generic identifiers, a method that differs from the ``\texttt{Lang}'' variant by mapping class names of each style to a fixed set of generic names (e.g., Style A, Style B). This approach improves model generalization over ``\texttt{Lang}'',  but generally falls short of the performance achieved with our optimal randomized identifiers ``\texttt{Lang, Rw}''. }}
    \label{fig:all_lp_performance}
\end{figure*}
\section{Results and Analysis}
\label{experimental_results}
We report macro-average F1 for style classification tasks following the XSLUE benchmark \cite{kang2021xslue}.  Our experimental results show that lexicon-based instructions can improve the zero-shot style classification performance in all settings, especially when source style meta-tuning and class randomization are involved.

\subsection{Pre-trained Language Models}
We experiment with T5 \cite{2020t5}, GPT-J \cite{gpt-j}, and LLaMA-2 \cite{touvron2023llama}. 
We also include experiments with the instruction-tuned models Flan-T5 \cite{flant5}  
and LLaMA-2-Chat \cite{touvron2023llama}, as these models have demonstrated the ability to effectively respond to instructions and generalize well to unseen tasks \cite{flant5,touvron2023llama}.\footnote{We ensure the evaluation style datasets are excluded from the fine-tuning tasks of Flan-T5, but it is unclear if LLaMA-2-Chat has been trained on these evaluation styles before.
} 
Implementation details are described in Appendix \ref{appendix:implementation_details}.

\subsection{Zero-shot Learning Results}
\label{main_experimental_results}
Table \ref{tabs:main_exp} shows zero-shot learning results for different methods on various  models.  We compare them with four distinct baseline methods, which are described in the caption of Table \ref{tabs:main_exp_baseline}.

\paragraph{Lexicon-based instructions outperform standard instructions.} In the zero-shot setup, incorporating lexicons into instructions demonstrates a significant advantage over the standard instructions without lexicon information across all the experimented models.
For example, after integrating lexicons into instructions and randomizing classes, {\LexPTmethod} improves upon the standard instructions by an average of 23.58 F1 points on Style-T5 and an average of 13.22 F1 points on Style-LLaMA. One possible explanation for this improvement is that fixing class names during source fine-tuning may lead the model to memorize these names, rather than meta-learning to classify target styles zero-shot. This is not ideal as our goal is to predict unseen styles and thus learning to use lexicons is important. By randomizing class names during meta-tuning, the model is forced to use lexicons, rather than relying on source-style identifiers (e.g., class names). 
Further experiments on class randomization are presented later in this section.

Furthermore, we find that lexicons seem to improve zero-shot style classification even without meta-tuning.  For example, by simply integrating lexicons into instructions, LLaMA-2-Chat models improve their performance in most styles. Notably, F1 jumps from 43.84 to 51.01 for Humor style on LLaMA-2-Chat (7B).

\paragraph{Meta-tuning on style data with lexicons enhances the zero-shot performance compared to general instruction tuning.} 
\label{format_transfer_analyses} Both Style-T5 and Style-LLaMA demonstrate a significant performance improvement over their general instruction-tuned counterparts, i.e., Flan-T5 and LLaMA-2-Chat, when lexicons are included.
For instance, Style-LLaMA (7B) outperforms LLaMA-2-Chat (7B) in five out of six styles, achieving an average increase of 4.28 F1 points. 
This suggests the benefits of lexicon-based instructions and the effectiveness of instruction tuning on training styles.

\paragraph{Class randomization matters in lexicon-based meta-tuning.} \label{lexicon-based-prompts}
\hspace{-0.06in}We study the impact of natural language descriptions and class randomization in our approach by independently fine-tuning Style-T5 on training styles using nine lexicon-based instruction variants (see \S \ref{variant_lps} below). Our experimental results in Figure \ref{fig:all_lp_performance} show that introducing class randomization can improve the zero-shot performance on the six unseen evaluation styles consistently. For example, the average F1 improves from 35.58 (\texttt{minimal}) to 50.54 (\texttt{R\#}).

\begin{table}[]
\centering
\small
\resizebox{\linewidth}{!}{%
\renewcommand{\arraystretch}{1.4}
\setlength{\tabcolsep}{1pt}
\begin{tabular}{l>{\centering\arraybackslash}p{1.8cm}>{\centering\arraybackslash}p{2cm}>{\centering\arraybackslash}p{1.5cm}>{\centering\arraybackslash}p{1.5cm}}
\toprule
     & \textbf{no rand.}            & \textbf{rand. indices}        & \multicolumn{2}{c}{\textbf{rand. words}}       \\ 
\midrule
\textbf{vocab size}  &  ---    & 3           & 3          & 18,843          \\ \hdashline[1pt/1pt]      
\textbf{w/o language} & minimal & R\# & Rw- & Rw \\
\hdashline[1pt/1pt]
\textbf{w/ language} & Lang & Lang, R\# & Lang,Rw- & Lang,Rw\\ 
\bottomrule
\end{tabular}}
\caption{Lexicon-based instruction variants \textcolor{black}{(as detailed in \S\ref{variant_lps})}. 
``vocab" is the fixed set of indices or words, from which a class name can be randomly selected. 
}\vskip -0.05in
\label{tabs:lp_differences}
\end{table}

\subsubsection{Details on Lexicon-based Instruction Variations with Class Randomization}
\label{variant_lps}
Here we describe the lexicon-based instruction formats tested, including different types of class randomization. All prompt variants are summarized in Table \ref{tabs:lp_differences}, while example prompts for each variant are shown in Figure \ref{fig:lp_examples} in the Appendix. ``R\#'' represents randomizing class names with numerical indices, and ``Rw'' means using random words as class names in the instruction. We simply use the default English word list in Ubuntu\footnote{\url{/usr/share/dict/words}} for this randomization. 
``Rw-" and ``Rw" differ only in the size of the fixed set from which a class name is sampled.
``Rw'' uses a much larger set (``vocab size'') compared to other variants, which reduces the chance of assigning the same word to the same class in different examples. 
This class randomization has pros and cons. On one hand, it may hurt performance because it prevents the model from inferring the meaning of classes from class names. On the other hand, it could enhance performance by encouraging the model to genuinely learn the input-class mappings and make use of lexicons, rather than memorizing class names from training styles that are observed during meta-training. Figure \ref{fig:all_lp_performance} shows that class randomization is helpful, possibly because the latter factor outweighs the former.

\subsection{Few-shot Learning Results}
\label{exp:few-shot-results}
Table \ref{tabs:few-shot-exp} shows results of few-shot learning methods.

\paragraph{Incorporating lexicons in few-shot learning reduces the sensitivity to example selection.} 
Prior work has shown that the choice of examples selected for few-shot learning can lead to very different performance \cite{Zhao2021CalibrateBU, liu-etal-2022-makes}. Hence how to reduce the sensitivity due to example selection has become an important question. In Table \ref{tabs:few-shot-exp}, we observe that after introducing lexicons into prompts, the standard deviation of performance across five runs generally decreases. For example, MetaICL$_4$ performs extremely unreliably on Romance with a high standard deviation of 27.30, while MetaICL$_4$+Lex not only improves performance but also stabilizes inference with a standard deviation dropped to 6.52. This suggests using lexicons may reduce a model's reliance on the selection of few-shot examples \citep{liu-etal-2022-makes}. 


\paragraph{Introducing lexicons into in-context examples can be beneficial when gold labels are absent.} 
When the examples of the evaluation style are randomly labeled \cite{2022RethinkingL}, introducing lexicons into MetaICL is generally more useful than increasing the number of examples. 
\textcolor{black}{In four out of six styles, MetaICL$_{16}$ shows a lower average across five runs, compared to MetaICL$_4$. The average F1 score (38.47) of MetaICL$_{16}$ over six styles is lower than that (42.79) of MetaICL$_4$. In contrast, MetaICL$_4$+Lex outperforms MetaICL$_4$, achieving an average F1 score of 45.29 over the six styles. Four styles show improved average performance across five runs with MetaICL$_4$+Lex, compared to MetaICL$_4$. For the other two styles (i.e., Shakespeare and Age) where MetaICL$_4$+Lex underperforms, their standard deviation in MetaICL$_4$+Lex is much smaller than in MetaICL$_4$.}
When ground-truth labels are accessible, MetaICL$_{16}$ showcases a superior average performance, suggesting that increasing the number of demonstrations might be more effective in this case.

\section{Generalization to Novel Styles}
\label{sec:gpt4-styles}
In prior sections, we demonstrated the effectiveness of lexicon meta-tuning on existing style datasets.  To demonstrate that our method is able to generalize beyond styles that have been previously studied in the NLP community, we  use LLMs to semi-automatically propose new styles, and then generate instances of text presenting each style (i.e., labeled examples).  The new styles generated in this section are then used to evaluate our approach's ability to generalize to styles that include but are not limited to niche literary genres, or rapidly evolving communication styles in social media (see examples in Table \ref{appendix:full_list_63_styles} and \ref{tab:case_study_set_example} in the Appendix).

\subsection{A Diverse Collection of New Styles}
\label{subsec:diverse_styles}

\paragraph{Style Creation.}
We compiled a diverse collection of language styles by initiating the data generation based on the 13 styles listed in Table \ref{tab:dataset_statistics}. This initial set served as seeds for prompting LLaMA-2-Chat 70B to generate new style classification tasks. We filtered out any LLM-generated tasks that did not align with our textual classification objective. To encourage diversity, a new task is added to the pool only when its ROUGE-L similarity with any existing task is less than 0.6. 
This process produced 58 new unique style classification tasks (full list in Appendix Table \ref{appendix:full_list_63_styles}). 
We then randomly divided these tasks into the training and evaluation split, avoiding task overlap. To further enrich the diversity, we developed and added 5 additional tasks to the evaluation split, such as composite chatbot styles (e.g., a blend of empathetic, colloquial, and humorous responses), and writing styles of various authors. Please refer to Appendix \ref{appendix:style_creation} for additional details on the style creation process.

\paragraph{Lexicon Creation.} 
To ensure consistency and clarity in our approach to style identification, we developed a lexicon for each new style. This was achieved by prompting LLaMA-2-Chat 70B, as detailed in Appendix \ref{appendix:lexicon_creation}, to generate a concise lexicon comprising up to five words or phrases for each style class. Depending on the construction method, these lexicons may vary in quality and size from a few words to thousands. Nevertheless, we demonstrate the benefits of our method with as few as five words per style sampled from lexicons (Appendix \ref{appendix:lexicon_size}), and highlight the robustness of lexicon-based instructions across various lexicon creation methods, particularly when class randomization is applied. 

\paragraph{Labeled Example Generation.}
We employed LLaMA-2-Chat 7B to generate 100 unique examples for each class in our training style split, 
which results in a training style dataset $\mathcal{D}_{\text{train}}$.
For the evaluation style dataset $\mathcal{D}_{\text{eval}}$, we used GPT-4 \cite{openai2023gpt4} to create high-quality stylistic examples. Through the OpenAI API, we generated 20 examples for each class at a total cost of \$9.11. Details about this process are in Appendix \ref{appendix:example_generation}, together with examples of lexicons in Table \ref{tab:case_study_set_example}.

\paragraph{Human Verification of Data Quality.} 
To measure the reliability of $\mathcal{D}_{\text{eval}}$, we asked three human annotators\footnote{The three annotators include: one of the authors, a graduate student in CS, and a mathematician.} to independently review a shared set of 500 randomly selected annotation examples. 
Annotators were instructed to assess the accuracy of labels for examples generated by GPT-4 and make necessary corrections, \textcolor{black}{as detailed in Appendix \ref{appendix:quality_check_eval_set}}. 
We computed inter-annotator agreement using Krippendorff's alpha. A high agreement score of 93.27\% reflects strong reliability in the annotations \cite{krippendorff2004reliability}. 

\begin{table}[h]
\centering
\small
\begin{adjustbox}{width=.95\columnwidth}
\begin{tabular}{lcc}
\toprule
statistics           &    $\mathcal{D}_{\text{train}}$  & $\mathcal{D}_{\text{eval}}$  \\ \midrule
\# of classification tasks         & 43  & 20  \\
\# of examples                     & 10,308 & 1,050  \\
avg. \# of classes per example & 3.20 & 3.12 \\
avg. example length (in words)  & 30.47 & 21.40 \\ 
avg. lexicon size (in words/phrases)   & 4.11 &3.74 \\ \bottomrule
\end{tabular}
\end{adjustbox}\vskip -0.03in
\caption{Statistics of model-generated datasets.}\vskip -0.07in
\label{tab:case_study_set_statistics}
\end{table}
\paragraph{Corpus Statistics.}
Our data generation process produced a collection of 11,358 distinctive examples, spanning 63 varied style classification tasks. 
Table \ref{tab:case_study_set_statistics} describes the statistics of our data. The distribution of $K$-class tasks (where $K$ is the number of distinct style classes to be distinguished) is illustrated in Figure \ref{fig:case_study_task_distribution} in the Appendix, showcasing the diverse range of styles included in our analysis.

\subsection{Experiments}
\paragraph{Experiment Setup.} 
We evaluated the zero-shot performance of LLaMA-2-Chat (7B, 13B) and Style-LLaMA (7B) on $\mathcal{D}_{\text{eval}}$. Given the balanced class distribution in this test set, we report accuracy in Table \ref{tab:case_study_results}. We also include Style-LLaMA+ (7B), which is a LLaMA-2 model fine-tuned on a mix of benchmark training styles and the training set $\mathcal{D}_{\text{train}}$ generated by LLaMA-2-Chat 7B. Note that the training set $\mathcal{D}_{\text{train}}$ and the evaluation set $\mathcal{D}_{\text{eval}}$ were created by different language models (\S \ref{subsec:diverse_styles}).
Implementation details are described in Appendix \ref{appendix:implementation_details}.  A baseline was set by randomly assigning a class to each example, averaging the results over five seeds. 
\paragraph{Results}
Table \ref{tab:case_study_results} demonstrates the advantages of lexicon-based instructions over the standard instructions. Notably, Style-LLaMA and Style-LLaMA+ show the most significant performance gains, with an average improvement of 12.46 and 8.85, respectively. This is likely because lexicon-based instruction-tuning enhances their adaptability to new styles through more effective lexicon usage. 
Furthermore, Style-LLaMA+ shows a substantial improvement over other models, suggesting that the inclusion of a diverse set of model-generated style training data can effectively enhance performance. 
The high score of Style-LLaMA+ with lexicon integration suggests that the combination of additional training data and lexicon-based instructions might be the most effective approach for generalization among the evaluated methods.

\begin{table}[]
\centering
\small
\begin{tabular}{lcc}
\toprule
 & \noLexMethod & {\LexPTmethod} \\ \midrule
Random Baseline     & \multicolumn{2}{c}{36.65} \\
\hdashline[1pt/1pt]
LLaMA-2-Chat (7B) & 53.09  &  56.23     \\
\emph{Style-LLaMA} (7B)  &  46.25  &  58.71 \\ 
\emph{Style-LLaMA+} (7B) & 65.46 & \textbf{74.31} \\
LLaMA-2-Chat (13B) & 56.80   &  59.75     \\\bottomrule
\end{tabular}
\vskip -0.03in
\caption{Zero-shot learning performance (accuracy) on $\mathcal{D}_{\text{eval}}$. Lexicon-based instructions improve the zero-shot generalization capabilities of the studied models. 
}\vskip -0.07in
\label{tab:case_study_results}
\end{table}

\section{Related Work}
\paragraph{Style classification.} 
Research in NLP has studied various language styles. \citet{kang2021xslue} provided a benchmark for fully-supervised style classification that combines many existing datasets for style classification, such as formality \cite{formality2018}, sarcasm \cite{sarc},  Hate/Offense (i.e., toxicity) \cite{davidson2017automated}, politeness \cite{politeness2013}, and sentiment \cite{socher-etal-2013-recursive,wang2021detecting}. 
Other writing styles include but are not limited to readability (i.e., simplicity) \cite{cefr}, Shakespearean English \cite{xu2012paraphrasing}, subjectivity \cite{subjectivity}, and engagingness \cite{jin2020hooks}. 
Despite an extensive range of style classification tasks studied in prior research, zero-shot or cross-style classification is relatively underexplored \cite{puri2019zero}. 
In particular, much of the cross-style research thus far has focused on text generation tasks \cite{text-style-transfer-survey,zhou2023controlled}, rather than classification. In this study, we aim to address this gap in the literature by concentrating on zero-shot style classification across a collection of diverse styles.

\paragraph{Language model prompting.} 
Large language models, such as GPT-3 \cite{brown2020language}, exhibit impressive zero-shot learning abilities when conditioned on appropriate textual contexts, i.e., prompts, or natural language instructions. 
Since then, how to design appropriate prompts has become a popular line of research \cite{t0,flant5}. 
In this work, we propose to incorporate lexicons into instructions and teach the model to better utilize stylistic lexicon knowledge through instruction tuning. Recently, \citet{zhou2023controlled} specified styles in instructions as constraints to improve controlled text generation. Parallel to our study, \citet{gao2023benefits} investigated label descriptions to enhance zero-shot topic and sentiment classification.
We focus on style classification, a challenging area in NLP characterized by its extensive scope and complexity, encompassing a wide range of stylistic expression across various domains of text. 
In order to improve the generalization capabilities of instruction-tuned models, we replace class names in instructions with entirely random words during fine-tuning on training styles. This is similar to \citet{dist3}, which indexes and shuffles slot descriptions in prompts for dialogue state tracking. Moreover, our work differs from the standard practice in previous studies \cite{2022RethinkingL, dist3, wei2023larger}, where a pre-defined set of class names, is equal in size to the number of labels in the associated datasets.

\section{Conclusion \& Discussion}
In this work, we study zero-shot style classification using large language models in combination with lexicon-based instructions. Experiments show that conventional instructions often struggle to generalize across diverse styles. However, our lexicon-based instruction  approach demonstrates the potential to fine-tune models for improved zero-shot generalization to unseen styles. 
By utilizing a diverse range of data sources, such as  Wikipedia, captions, social media (Table \ref{tab:dataset_statistics}), and LLM-generated data (\S \ref{sec:gpt4-styles}), we ensure the robustness and generalizability of our findings across various contexts. 

Furthermore, the wide range of tasks, together with the flexible and effective usage of lexicons highlights potential applications of our approach. 
For example, our approach can be used to identify harmful or toxic content, interpret the implicit meaning in communication (e.g., humor, sarcasm, friendliness, hostility, etc.), and assess text readability. More potential applications can be found in Appendix Table \ref{appendix:full_list_63_styles}.  In addition, our method may generalize to generation tasks (e.g., text style transfer), which we intend to explore in future work.

\section*{Limitations}
\textcolor{black}{The ambiguity of style in language often poses a challenge for classification efforts. Individual interpretations of styles can vary widely based on personal, cultural, and contextual factors \cite{hovy1987generating}, underscoring the nuanced nature of language and the difficulty of categorizing it into discrete styles. Our method, which utilizes lexicon-based instruction, mitigates this issue by focusing on stylistically expressive words or phrases rather than relying solely on style names. This approach offers a degree of flexibility in representing styles, although it does not fully resolve the underlying complexity. Future research should continue to explore these nuances, striving for definitions that are both precise and adaptable to the diverse ways in which style manifests.}

Additionally, in our method, we leverage the lexicons we have collected (as detailed in Table \ref{tab:dataset_statistics}). However, it is important to acknowledge that a potential limitation of our approach lies in the possibility of different performance outcomes when using lexicons of varying qualities. While we have conducted comparisons between lexicons from different sources in Appendix \ref{ablate:lex_source}, it is plausible that utilizing different lexicons could yield different results. 

\textcolor{black}{Another limitation is that we only include a limited set of styles in English for evaluation in \S \ref{experimental_results}, due to availability of high-quality style datasets and lexicons. While we attempted to assess the generalization capabilities of our method to a novel set of GPT-4 generated language styles in \S\ref{sec:gpt4-styles}, it is important to recognize that they might not be representative of the styles that users wish to analyze using our approach. 
Moreover, our artificially balanced datasets may not accurately reflect the imbalanced label distribution in real-world scenarios.  
Furthermore, utilizing LLMs to generate datasets can perpetuate the inherent biases within LLMs and introduce biases through the choice of prompts used \cite{wang-etal-2023-self-instruct}, which complicates the task of creating unbiased and representative datasets.
Future work could focus on curating human-written datasets to assess robustness and applicability of lexicon meta-tuning in practical applications.}

\section*{Ethical Considerations} Style classification is widely studied in the NLP research community. We strictly limit to using only the existing and commonly used datasets that are related to demographic information in our experiments. As a proof of concept, this research study was only conducted on English data, where human annotations for multiple styles are available for use in the evaluation. We also acknowledge that linguistic styles are not limited to what are included in this paper, and can be much more diverse. Future efforts in the NLP community could further extend research on stylistics to more languages and styles.

\section*{Acknowledgements}
We thank Tarek Naous, Michael Ryan, Fan Bai, Shuheng Liu, Junmo Kang, Duong Minh Le and anonymous reviewers for their helpful feedback on this work.
We also would like to thank Azure’s Accelerate Foundation Models Research Program graciously providing access to API-based models, such as GPT-4.
This research is supported in part by the NSF (IIS-2052498, IIS-2144493 and  IIS-2112633), ODNI and IARPA via the HIATUS program (2022-22072200004). The views and conclusions contained herein are those of the authors and should not be interpreted as necessarily representing the official policies, either expressed or implied, of NSF, ODNI, IARPA, or the U.S. Government. The U.S. Government is authorized to reproduce and distribute reprints for governmental purposes notwithstanding any copyright annotation therein.

\bibliography{anthology,custom}

\begin{thebibliography}{56}
\expandafter\ifx\csname natexlab\endcsname\relax\def\natexlab#1{#1}\fi

\bibitem[{Ahn(2005)}]{hateoffense-nlp-lexicon}
Luis~Von Ahn. 2005.
\newblock Useful resources: Offensive/profane word list.
\newblock \url{https://www.cs.cmu.edu/~biglou/resources/}.

\bibitem[{Arase et~al.(2022)Arase, Uchida, and Kajiwara}]{cefr}
Yuki Arase, Satoru Uchida, and Tomoyuki Kajiwara. 2022.
\newblock \href {https://aclanthology.org/2022.emnlp-main.416} {{CEFR}-based sentence difficulty annotation and assessment}.
\newblock In \emph{Proceedings of the 2022 Conference on Empirical Methods in Natural Language Processing}, pages 6206--6219, Abu Dhabi, United Arab Emirates. Association for Computational Linguistics.

\bibitem[{Artstein and Poesio(2008)}]{artstein2008inter}
Ron Artstein and Massimo Poesio. 2008.
\newblock Inter-coder agreement for computational linguistics.
\newblock \emph{Computational linguistics}, 34(4):555--596.

\bibitem[{Brown et~al.(2020)Brown, Mann, Ryder, Subbiah, Kaplan, Dhariwal, Neelakantan, Shyam, Sastry, Askell et~al.}]{brown2020language}
Tom Brown, Benjamin Mann, Nick Ryder, Melanie Subbiah, Jared~D Kaplan, Prafulla Dhariwal, Arvind Neelakantan, Pranav Shyam, Girish Sastry, Amanda Askell, et~al. 2020.
\newblock Language models are few-shot learners.
\newblock \emph{Advances in neural information processing systems}.

\bibitem[{Chen and Ritter(2021)}]{chen2021model}
Yang Chen and Alan Ritter. 2021.
\newblock Model selection for cross-lingual transfer.
\newblock In \emph{Proceedings of the 2021 Conference on Empirical Methods in Natural Language Processing}.

\bibitem[{Chung et~al.(2022)Chung, Hou, Longpre, Zoph, Tay, Fedus, Li, Wang, Dehghani, Brahma, Webson, Gu, Dai, Suzgun, Chen, Chowdhery, Castro-Ros, Pellat, Robinson, Valter, Narang, Mishra, Yu, Zhao, Huang, Dai, Yu, Petrov, Chi, Dean, Devlin, Roberts, Zhou, Le, and Wei}]{flant5}
Hyung~Won Chung, Le~Hou, Shayne Longpre, Barret Zoph, Yi~Tay, William Fedus, Yunxuan Li, Xuezhi Wang, Mostafa Dehghani, Siddhartha Brahma, Albert Webson, Shixiang~Shane Gu, Zhuyun Dai, Mirac Suzgun, Xinyun Chen, Aakanksha Chowdhery, Alex Castro-Ros, Marie Pellat, Kevin Robinson, Dasha Valter, Sharan Narang, Gaurav Mishra, Adams Yu, Vincent Zhao, Yanping Huang, Andrew Dai, Hongkun Yu, Slav Petrov, Ed~H. Chi, Jeff Dean, Jacob Devlin, Adam Roberts, Denny Zhou, Quoc~V. Le, and Jason Wei. 2022.
\newblock \href {https://doi.org/10.48550/ARXIV.2210.11416} {Scaling instruction-finetuned language models}.

\bibitem[{CrowdTruth(2016)}]{shorthumor}
CrowdTruth. 2016.
\newblock {Short Text Corpus For Humor Detection}.
\newblock \url{http://github.com/CrowdTruth/Short-Text-Corpus-For-Humor-Detection}.
\newblock [Online; accessed 1-Oct-2019].

\bibitem[{Danescu-Niculescu-Mizil et~al.(2013)Danescu-Niculescu-Mizil, Sudhof, Jurafsky, Leskovec, and Potts}]{politeness2013}
Cristian Danescu-Niculescu-Mizil, Moritz Sudhof, Dan Jurafsky, Jure Leskovec, and Christopher Potts. 2013.
\newblock \href {https://www.aclweb.org/anthology/P13-1025} {A computational approach to politeness with application to social factors}.
\newblock In \emph{Proceedings of the 51st Annual Meeting of the Association for Computational Linguistics (Volume 1: Long Papers)}, pages 250--259, Sofia, Bulgaria. Association for Computational Linguistics.

\bibitem[{Davidson et~al.(2017)Davidson, Warmsley, Macy, and Weber}]{davidson2017automated}
Thomas Davidson, Dana Warmsley, Michael Macy, and Ingmar Weber. 2017.
\newblock Automated hate speech detection and the problem of offensive language.
\newblock In \emph{Proceedings of the 11th International AAAI Conference on Web and Social Media}, ICWSM '17, pages 512--515.

\bibitem[{Dettmers and Zettlemoyer(2023)}]{dettmers2023case}
Tim Dettmers and Luke Zettlemoyer. 2023.
\newblock The case for 4-bit precision: k-bit inference scaling laws.
\newblock In \emph{International Conference on Machine Learning}, pages 7750--7774. PMLR.

\bibitem[{Dodds et~al.(2015)Dodds, Clark, Desu, Frank, Reagan, Williams, Mitchell, Harris, Kloumann, Bagrow et~al.}]{dodds2015human}
Peter~Sheridan Dodds, Eric~M Clark, Suma Desu, Morgan~R Frank, Andrew~J Reagan, Jake~Ryland Williams, Lewis Mitchell, Kameron~Decker Harris, Isabel~M Kloumann, James~P Bagrow, et~al. 2015.
\newblock Human language reveals a universal positivity bias.
\newblock \emph{Proceedings of the national academy of sciences}, 112(8):2389--2394.

\bibitem[{Eisenstein(2017)}]{eisenstein2017unsupervised}
Jacob Eisenstein. 2017.
\newblock Unsupervised learning for lexicon-based classification.
\newblock In \emph{Proceedings of the AAAI Conference on Artificial Intelligence}, volume~31.

\bibitem[{Gao et~al.(2023)Gao, Ghosh, and Gimpel}]{gao2023benefits}
Lingyu Gao, Debanjan Ghosh, and Kevin Gimpel. 2023.
\newblock The benefits of label-description training for zero-shot text classification.
\newblock \emph{arXiv preprint arXiv:2305.02239}.

\bibitem[{Hovy(1987)}]{hovy1987generating}
Eduard Hovy. 1987.
\newblock Generating natural language under pragmatic constraints.
\newblock \emph{Journal of Pragmatics}, 11(6):689--719.

\bibitem[{Hu et~al.(2021)Hu, Shen, Wallis, Allen-Zhu, Li, Wang, Wang, and Chen}]{hu2021lora}
Edward~J Hu, Yelong Shen, Phillip Wallis, Zeyuan Allen-Zhu, Yuanzhi Li, Shean Wang, Lu~Wang, and Weizhu Chen. 2021.
\newblock Lora: Low-rank adaptation of large language models.
\newblock \emph{arXiv preprint arXiv:2106.09685}.

\bibitem[{Jin et~al.(2022)Jin, Jin, Hu, Vechtomova, and Mihalcea}]{text-style-transfer-survey}
Di~Jin, Zhijing Jin, Zhiting Hu, Olga Vechtomova, and Rada Mihalcea. 2022.
\newblock \href {https://doi.org/10.1162/coli_a_00426} {{Deep Learning for Text Style Transfer: A Survey}}.
\newblock \emph{Computational Linguistics}, 48(1):155--205.

\bibitem[{Jin et~al.(2020)Jin, Jin, Zhou, Orii, and Szolovits}]{jin2020hooks}
Di~Jin, Zhijing Jin, Joey~Tianyi Zhou, Lisa Orii, and Peter Szolovits. 2020.
\newblock Hooks in the headline: Learning to generate headlines with controlled styles.
\newblock \emph{arXiv preprint arXiv:2004.01980}.

\bibitem[{Kang and Hovy(2021)}]{kang2021xslue}
Dongyeop Kang and Eduard Hovy. 2021.
\newblock Style is not a single variable: Case studies for cross-style language understanding.
\newblock In \emph{Proceedings of the 59th Annual Meeting of the Association for Computational Linguistics}. Association for Computational Linguistics.

\bibitem[{Kang et~al.(2023)Kang, Luo, Zhu, Glass, Cox, Ritter, Feris, and Karlinsky}]{kang2023self}
Junmo Kang, Hongyin Luo, Yada Zhu, James Glass, David Cox, Alan Ritter, Rogerio Feris, and Leonid Karlinsky. 2023.
\newblock Self-specialization: Uncovering latent expertise within large language models.
\newblock \emph{arXiv preprint arXiv:2310.00160}.

\bibitem[{Khodak et~al.(2018)Khodak, Saunshi, and Vodrahalli}]{sarc}
Mikhail Khodak, Nikunj Saunshi, and Kiran Vodrahalli. 2018.
\newblock \href {https://aclanthology.org/L18-1102} {A large self-annotated corpus for sarcasm}.
\newblock In \emph{Proceedings of the Eleventh International Conference on Language Resources and Evaluation ({LREC} 2018)}, Miyazaki, Japan. European Language Resources Association (ELRA).

\bibitem[{Krippendorff(2004)}]{krippendorff2004reliability}
Klaus Krippendorff. 2004.
\newblock Reliability in content analysis: Some common misconceptions and recommendations.
\newblock \emph{Human communication research}, 30(3):411--433.

\bibitem[{Liu et~al.(2022)Liu, Shen, Zhang, Dolan, Carin, and Chen}]{liu-etal-2022-makes}
Jiachang Liu, Dinghan Shen, Yizhe Zhang, Bill Dolan, Lawrence Carin, and Weizhu Chen. 2022.
\newblock \href {https://doi.org/10.18653/v1/2022.deelio-1.10} {What makes good in-context examples for {GPT}-3?}
\newblock In \emph{Proceedings of Deep Learning Inside Out (DeeLIO 2022): The 3rd Workshop on Knowledge Extraction and Integration for Deep Learning Architectures}, pages 100--114, Dublin, Ireland and Online. Association for Computational Linguistics.

\bibitem[{Maddela and Xu(2018)}]{maddela-xu-2018-word}
Mounica Maddela and Wei Xu. 2018.
\newblock \href {https://doi.org/10.18653/v1/D18-1410} {A word-complexity lexicon and a neural readability ranking model for lexical simplification}.
\newblock In \emph{Proceedings of the 2018 Conference on Empirical Methods in Natural Language Processing}, pages 3749--3760, Brussels, Belgium. Association for Computational Linguistics.

\bibitem[{Meng et~al.(2020)Meng, Zhang, Huang, Xiong, Ji, Zhang, and Han}]{2020textUseLabelNamesOnly}
Yu~Meng, Yunyi Zhang, Jiaxin Huang, Chenyan Xiong, Heng Ji, Chao Zhang, and Jiawei Han. 2020.
\newblock Text classification using label names only: A language model self-training approach.
\newblock In \emph{Proceedings of the 2020 Conference on Empirical Methods in Natural Language Processing}.

\bibitem[{Min et~al.(2022{\natexlab{a}})Min, Lewis, Zettlemoyer, and Hajishirzi}]{min-etal-2022-metaicl}
Sewon Min, Mike Lewis, Luke Zettlemoyer, and Hannaneh Hajishirzi. 2022{\natexlab{a}}.
\newblock \href {https://doi.org/10.18653/v1/2022.naacl-main.201} {{M}eta{ICL}: Learning to learn in context}.
\newblock In \emph{Proceedings of the 2022 Conference of the North American Chapter of the Association for Computational Linguistics: Human Language Technologies}, pages 2791--2809, Seattle, United States. Association for Computational Linguistics.

\bibitem[{Min et~al.(2022{\natexlab{b}})Min, Lyu, Holtzman, Artetxe, Lewis, Hajishirzi, and Zettlemoyer}]{2022RethinkingL}
Sewon Min, Xinxi Lyu, Ari Holtzman, Mikel Artetxe, Mike Lewis, Hannaneh Hajishirzi, and Luke Zettlemoyer. 2022{\natexlab{b}}.
\newblock Rethinking the role of demonstrations: What makes in-context learning work?
\newblock In \emph{EMNLP}.

\bibitem[{Mohammad et~al.(2013)Mohammad, Kiritchenko, and Zhu}]{mohammad2013nrc}
Saif Mohammad, Svetlana Kiritchenko, and Xiaodan Zhu. 2013.
\newblock Nrc-canada: Building the state-of-the-art in sentiment analysis of tweets.
\newblock In \emph{Second Joint Conference on Lexical and Computational Semantics (* SEM), Volume 2: Proceedings of the Seventh International Workshop on Semantic Evaluation (SemEval 2013)}.

\bibitem[{Mohammad and Turney(2010)}]{mohammad2010emotions}
Saif Mohammad and Peter Turney. 2010.
\newblock Emotions evoked by common words and phrases: Using mechanical turk to create an emotion lexicon.
\newblock In \emph{Proceedings of the NAACL HLT 2010 workshop on computational approaches to analysis and generation of emotion in text}.

\bibitem[{Mohammad(2021)}]{SentimentEmotionSurvey2021}
Saif~M. Mohammad. 2021.
\newblock Sentiment analysis: Automatically detecting valence, emotions, and other affectual states from text.
\newblock In Herb Meiselman, editor, \emph{Emotion Measurement (Second Edition)}. Elsevier.

\bibitem[{OpenAI(2023)}]{openai2023gpt4}
OpenAI. 2023.
\newblock \href {http://arxiv.org/abs/2303.08774} {Gpt-4 technical report}.

\bibitem[{Pang and Lee(2004)}]{subjectivity}
Bo~Pang and Lillian Lee. 2004.
\newblock A sentimental education: Sentiment analysis using subjectivity summarization based on minimum cuts.
\newblock In \emph{Annual Meeting of the Association for Computational Linguistics}.

\bibitem[{Paszke et~al.(2019)Paszke, Gross, Massa, Lerer, Bradbury, Chanan, Killeen, Lin, Gimelshein, Antiga, Desmaison, Köpf, Yang, DeVito, Raison, Tejani, Chilamkurthy, Steiner, Fang, Bai, and Chintala}]{pytorch}
Adam Paszke, Sam Gross, Francisco Massa, Adam Lerer, James Bradbury, Gregory Chanan, Trevor Killeen, Zeming Lin, Natalia Gimelshein, Luca Antiga, Alban Desmaison, Andreas Köpf, Edward Yang, Zach DeVito, Martin Raison, Alykhan Tejani, Sasank Chilamkurthy, Benoit Steiner, Lu~Fang, Junjie Bai, and Soumith Chintala. 2019.
\newblock \href {https://doi.org/10.48550/ARXIV.1912.01703} {Pytorch: An imperative style, high-performance deep learning library}.

\bibitem[{Puri and Catanzaro(2019)}]{puri2019zero}
Raul Puri and Bryan Catanzaro. 2019.
\newblock Zero-shot text classification with generative language models.
\newblock \emph{arXiv preprint arXiv:1912.10165}.

\bibitem[{Raffel et~al.(2020)Raffel, Shazeer, Roberts, Lee, Narang, Matena, Zhou, Li, and Liu}]{2020t5}
Colin Raffel, Noam Shazeer, Adam Roberts, Katherine Lee, Sharan Narang, Michael Matena, Yanqi Zhou, Wei Li, and Peter~J. Liu. 2020.
\newblock \href {http://jmlr.org/papers/v21/20-074.html} {Exploring the limits of transfer learning with a unified text-to-text transformer}.
\newblock \emph{Journal of Machine Learning Research}, 21(140):1--67.

\bibitem[{Rao and Tetreault(2018)}]{formality2018}
Sudha Rao and Joel Tetreault. 2018.
\newblock \href {https://doi.org/10.18653/v1/N18-1012} {Dear sir or madam, may {I} introduce the {GYAFC} dataset: Corpus, benchmarks and metrics for formality style transfer}.
\newblock In \emph{Proceedings of the 2018 Conference of the North {A}merican Chapter of the Association for Computational Linguistics: Human Language Technologies, Volume 1 (Long Papers)}, pages 129--140, New Orleans, Louisiana. Association for Computational Linguistics.

\bibitem[{Sanh et~al.(2021)Sanh, Webson, Raffel, Bach, Sutawika, Alyafeai, Chaffin, Stiegler, Scao, Raja, Dey, Bari, Xu, Thakker, Sharma, Szczechla, Kim, Chhablani, Nayak, Datta, Chang, Jiang, Wang, Manica, Shen, Yong, Pandey, Bawden, Wang, Neeraj, Rozen, Sharma, Santilli, F{\'{e}}vry, Fries, Teehan, Biderman, Gao, Bers, Wolf, and Rush}]{t0}
Victor Sanh, Albert Webson, Colin Raffel, Stephen~H. Bach, Lintang Sutawika, Zaid Alyafeai, Antoine Chaffin, Arnaud Stiegler, Teven~Le Scao, Arun Raja, Manan Dey, M.~Saiful Bari, Canwen Xu, Urmish Thakker, Shanya Sharma, Eliza Szczechla, Taewoon Kim, Gunjan Chhablani, Nihal~V. Nayak, Debajyoti Datta, Jonathan Chang, Mike~Tian{-}Jian Jiang, Han Wang, Matteo Manica, Sheng Shen, Zheng~Xin Yong, Harshit Pandey, Rachel Bawden, Thomas Wang, Trishala Neeraj, Jos Rozen, Abheesht Sharma, Andrea Santilli, Thibault F{\'{e}}vry, Jason~Alan Fries, Ryan Teehan, Stella Biderman, Leo Gao, Tali Bers, Thomas Wolf, and Alexander~M. Rush. 2021.
\newblock \href {http://arxiv.org/abs/2110.08207} {Multitask prompted training enables zero-shot task generalization}.
\newblock \emph{CoRR}, abs/2110.08207.

\bibitem[{Sim et~al.(2013)Sim, Acree, Gross, and Smith}]{politics_public_lex}
Yanchuan Sim, Brice D.~L. Acree, Justin~H. Gross, and Noah~A. Smith. 2013.
\newblock \href {https://aclanthology.org/D13-1010} {Measuring ideological proportions in political speeches}.
\newblock In \emph{Proceedings of the 2013 Conference on Empirical Methods in Natural Language Processing}, pages 91--101, Seattle, Washington, USA. Association for Computational Linguistics.

\bibitem[{Socher et~al.(2013)Socher, Perelygin, Wu, Chuang, Manning, Ng, and Potts}]{socher-etal-2013-recursive}
Richard Socher, Alex Perelygin, Jean Wu, Jason Chuang, Christopher~D. Manning, Andrew Ng, and Christopher Potts. 2013.
\newblock \href {https://aclanthology.org/D13-1170} {Recursive deep models for semantic compositionality over a sentiment treebank}.
\newblock In \emph{Proceedings of the 2013 Conference on Empirical Methods in Natural Language Processing}, pages 1631--1642, Seattle, Washington, USA. Association for Computational Linguistics.

\bibitem[{Taboada et~al.(2011)Taboada, Brooke, Tofiloski, Voll, and Stede}]{taboada2011lexicon}
Maite Taboada, Julian Brooke, Milan Tofiloski, Kimberly Voll, and Manfred Stede. 2011.
\newblock Lexicon-based methods for sentiment analysis.
\newblock \emph{Computational linguistics}.

\bibitem[{Tausczik and Pennebaker(2010)}]{tausczik2010psychological}
Yla~R Tausczik and James~W Pennebaker. 2010.
\newblock The psychological meaning of words: Liwc and computerized text analysis methods.
\newblock \emph{Journal of language and social psychology}.

\bibitem[{Teng et~al.(2016)Teng, Vo, and Zhang}]{teng-etal-2016-context}
Zhiyang Teng, Duy-Tin Vo, and Yue Zhang. 2016.
\newblock \href {https://doi.org/10.18653/v1/D16-1169} {Context-sensitive lexicon features for neural sentiment analysis}.
\newblock In \emph{Proceedings of the 2016 Conference on Empirical Methods in Natural Language Processing}, pages 1629--1638, Austin, Texas. Association for Computational Linguistics.

\bibitem[{Touvron et~al.(2023)Touvron, Martin, Stone, Albert, Almahairi, Babaei, Bashlykov, Batra, Bhargava, Bhosale et~al.}]{touvron2023llama}
Hugo Touvron, Louis Martin, Kevin Stone, Peter Albert, Amjad Almahairi, Yasmine Babaei, Nikolay Bashlykov, Soumya Batra, Prajjwal Bhargava, Shruti Bhosale, et~al. 2023.
\newblock Llama 2: Open foundation and fine-tuned chat models.
\newblock \emph{arXiv preprint arXiv:2307.09288}.

\bibitem[{Verma and Srinivasan(2019)}]{verma2019lexical}
Gaurav Verma and Balaji~Vasan Srinivasan. 2019.
\newblock A lexical, syntactic, and semantic perspective for understanding style in text.
\newblock \emph{arXiv preprint arXiv:1909.08349}.

\bibitem[{Wang and Komatsuzaki(2021)}]{gpt-j}
Ben Wang and Aran Komatsuzaki. 2021.
\newblock {GPT-J-6B: A 6 Billion Parameter Autoregressive Language Model}.
\newblock \url{https://github.com/kingoflolz/mesh-transformer-jax}.

\bibitem[{Wang et~al.(2021)Wang, Lv, Mazumder, and Liu}]{wang2021detecting}
Shuai Wang, Guangyi Lv, Sahisnu Mazumder, and Bing Liu. 2021.
\newblock Detecting domain polarity-changes of words in a sentiment lexicon.
\newblock In \emph{Findings of the Association for Computational Linguistics: ACL-IJCNLP 2021}, pages 3657--3668.

\bibitem[{Wang et~al.(2010)Wang, Brooke, and Hirst}]{formalityLexicon2010}
Tong Wang, Julian Brooke, and Graeme Hirst. 2010.
\newblock \href {https://api.semanticscholar.org/CorpusID:6742665} {Inducing lexicons of formality from corpora}.

\bibitem[{Wang et~al.(2023)Wang, Kordi, Mishra, Liu, Smith, Khashabi, and Hajishirzi}]{wang-etal-2023-self-instruct}
Yizhong Wang, Yeganeh Kordi, Swaroop Mishra, Alisa Liu, Noah~A. Smith, Daniel Khashabi, and Hannaneh Hajishirzi. 2023.
\newblock \href {https://doi.org/10.18653/v1/2023.acl-long.754} {Self-instruct: Aligning language models with self-generated instructions}.
\newblock In \emph{Proceedings of the 61st Annual Meeting of the Association for Computational Linguistics (Volume 1: Long Papers)}, pages 13484--13508, Toronto, Canada. Association for Computational Linguistics.

\bibitem[{Wei et~al.(2023)Wei, Wei, Tay, Tran, Webson, Lu, Chen, Liu, Huang, Zhou et~al.}]{wei2023larger}
Jerry Wei, Jason Wei, Yi~Tay, Dustin Tran, Albert Webson, Yifeng Lu, Xinyun Chen, Hanxiao Liu, Da~Huang, Denny Zhou, et~al. 2023.
\newblock Larger language models do in-context learning differently.
\newblock \emph{arXiv preprint arXiv:2303.03846}.

\bibitem[{Wilson et~al.(2005)Wilson, Wiebe, and Hoffmann}]{wilson-etal-2005-recognizing}
Theresa Wilson, Janyce Wiebe, and Paul Hoffmann. 2005.
\newblock \href {https://aclanthology.org/H05-1044} {Recognizing contextual polarity in phrase-level sentiment analysis}.
\newblock In \emph{Proceedings of Human Language Technology Conference and Conference on Empirical Methods in Natural Language Processing}, pages 347--354, Vancouver, British Columbia, Canada. Association for Computational Linguistics.

\bibitem[{Wolf et~al.(2020)Wolf, Debut, Sanh, Chaumond, Delangue, Moi, Cistac, Rault, Louf, Funtowicz, Davison, Shleifer, von Platen, Ma, Jernite, Plu, Xu, Le~Scao, Gugger, Drame, Lhoest, and Rush}]{wolf-etal-2020-transformers}
Thomas Wolf, Lysandre Debut, Victor Sanh, Julien Chaumond, Clement Delangue, Anthony Moi, Pierric Cistac, Tim Rault, Remi Louf, Morgan Funtowicz, Joe Davison, Sam Shleifer, Patrick von Platen, Clara Ma, Yacine Jernite, Julien Plu, Canwen Xu, Teven Le~Scao, Sylvain Gugger, Mariama Drame, Quentin Lhoest, and Alexander Rush. 2020.
\newblock \href {https://doi.org/10.18653/v1/2020.emnlp-demos.6} {Transformers: State-of-the-art natural language processing}.
\newblock In \emph{Proceedings of the 2020 Conference on Empirical Methods in Natural Language Processing: System Demonstrations}, pages 38--45, Online. Association for Computational Linguistics.

\bibitem[{Xu(2017)}]{xu-2017-shakespeare}
Wei Xu. 2017.
\newblock \href {https://doi.org/10.18653/v1/W17-4901} {From shakespeare to {T}witter: What are language styles all about?}
\newblock In \emph{Proceedings of the Workshop on Stylistic Variation}, pages 1--9, Copenhagen, Denmark. Association for Computational Linguistics.

\bibitem[{Xu et~al.(2012)Xu, Ritter, Dolan, Grishman, and Cherry}]{xu2012paraphrasing}
Wei Xu, Alan Ritter, Bill Dolan, Ralph Grishman, and Colin Cherry. 2012.
\newblock Paraphrasing for style.
\newblock In \emph{COLING}, pages 2899--2914.

\bibitem[{Zhao et~al.(2022)Zhao, Gupta, Cao, Yu, Wang, Lee, Rastogi, Shafran, and Wu}]{dist3}
Jeffrey Zhao, Raghav Gupta, Yuan Cao, Dian Yu, Mingqiu Wang, Harrison Lee, Abhinav Rastogi, Izhak Shafran, and Yonghui Wu. 2022.
\newblock \href {https://doi.org/10.48550/ARXIV.2201.08904} {Description-driven task-oriented dialog modeling}.

\bibitem[{Zhao et~al.(2021)Zhao, Wallace, Feng, Klein, and Singh}]{Zhao2021CalibrateBU}
Tony Zhao, Eric Wallace, Shi Feng, Dan Klein, and Sameer Singh. 2021.
\newblock Calibrate before use: Improving few-shot performance of language models.
\newblock \emph{ArXiv}, abs/2102.09690.

\bibitem[{Zhong et~al.(2021)Zhong, Lee, Zhang, and Klein}]{zhong2021adapting}
Ruiqi Zhong, Kristy Lee, Zheng Zhang, and Dan Klein. 2021.
\newblock Adapting language models for zero-shot learning by meta-tuning on dataset and prompt collections.
\newblock In \emph{Findings of the Association for Computational Linguistics: EMNLP 2021}, pages 2856--2878.

\bibitem[{Zhou et~al.(2023)Zhou, Jiang, Wilcox, Cotterell, and Sachan}]{zhou2023controlled}
Wangchunshu Zhou, Yuchen~Eleanor Jiang, Ethan Wilcox, Ryan Cotterell, and Mrinmaya Sachan. 2023.
\newblock Controlled text generation with natural language instructions.
\newblock \emph{arXiv preprint arXiv:2304.14293}.

\end{thebibliography}
\bibliographystyle{acl_natbib}

\clearpage
\newpage
\appendix

\section{Benchmark Datasets Details}
\label{dataset_selection_from_xslue}
The XSLUE benchmark, designed for exploring cross-style language understanding, encompasses 15 styles \cite{kang2021xslue}. We choose 10 writing styles from XSLUE based on their suitability for our task. Specifically, we consider the task type (i.e., whether the task is classification or not), task granularity (e.g., whether the annotated style is sentence-level or not), expressiveness at both the word and phrase level (i.e., the possibility of expressing a style with lexicons). For example, the TroFi dataset for style Metaphor is not used because it is focused on the literal usage of one specific verb in a sentence. Take the verb ``drink" as an example, it is a literal expression in the sentence \say{\say{I stayed home and drank for two years after that,} he notes sadly}, whereas in \say{So the kids gave Mom a watch, said a couple of nice things, and drank a retirement toast in her honor}, \say{drink} is non-literal.

\section{Benchmark Lexicons Details}
\label{appendix:lexicon_all_details}
\subsection{Lexicon Creation}
\begin{figure*}
     \centering
     \begin{subfigure}[b]{0.5\textwidth}
         \centering
         \includegraphics[width=\textwidth]{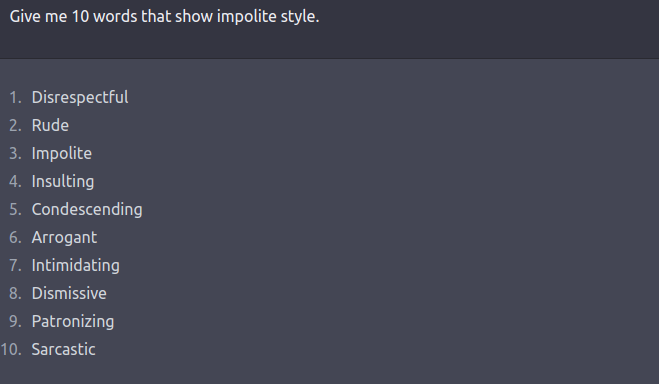}
         \caption{impolite}
         \label{fig:impolite_chatgpt}
     \end{subfigure}
     \begin{subfigure}[b]{0.4\textwidth}
         \centering
         \includegraphics[width=\textwidth]{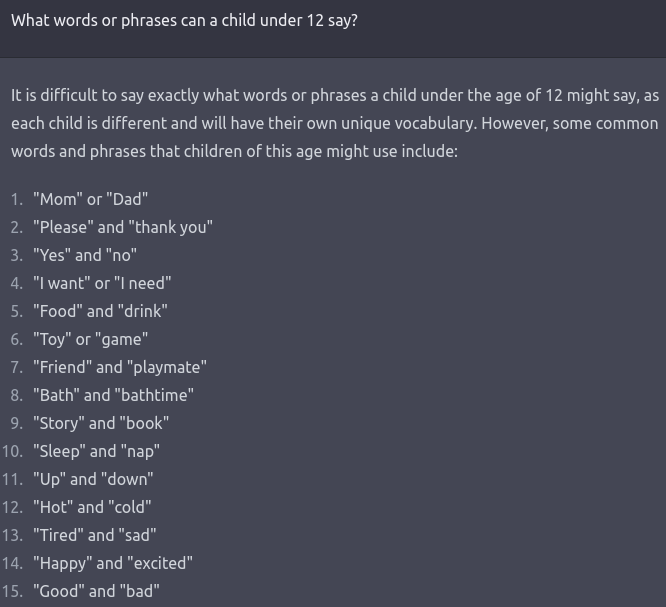}
         \caption{under12}
         \label{fig:under12_chatgpt}
     \end{subfigure}
        \caption{Examples of ChatGPT output for different style classes.}
        \label{fig:example_chatgpt_output}
\end{figure*}

\paragraph{ChatGPT-generated Lexicons.} Prior work has used models, such as BERT to generate class vocabularies for topic classification \cite{2020textUseLabelNamesOnly}. Inspired by this approach, we utilize the knowledge of LLMs by prompting them to generate a list of words that express the specific class of a style. In a preliminary study, we experimented with many LLMs, including BERT, GPT-J, GPT-NeoX, GPT-3.5 and ChatGPT. Among all, ChatGPT performs the best, so we use it to generate the lexicons.
Table \ref{tabs:chatgpt_prompt} shows the prompts we used for ChatGPT. Figure \ref{fig:example_chatgpt_output} presents some examples of ChatGPT output. 

\paragraph{Dictionary-based Lexicons.} We also considered lexicons generated by extracting the definition of each style from Google Dictionary.

\begin{table*}
\centering
\small
\begin{tabular}{lll}
\toprule
\textbf{Style}    & \textbf{Class}    & \multicolumn{1}{c}{\textbf{ChatGPT Prompt}} \\ \midrule
Politeness               & impolite                                                  & \begin{tabular}[c]{@{}l@{}}Give me 10 words that show impolite style.\\ Give me 20 words or short phrases that people may use when they show impolite \\ attitude towards others.\end{tabular}                                                                                                \\ \midrule
Romance                  & literal                                                   & \begin{tabular}[c]{@{}l@{}}What's the difference between literal text and romantic text?\\ Give me 20 words or short phrases that show the literal style rather than romantic\\ style.\end{tabular}                                                                                           \\ \midrule
\multirow{2}{*}{Humor}   & humorous                                                  & \begin{tabular}[c]{@{}l@{}}Give me 10 words that show humorous style.\\ Give me 20 words or short phrases that people may use in text to show humor.\end{tabular}                                                                                                                             \\
                         & literal                                                   & \begin{tabular}[c]{@{}l@{}}What's the difference between literal text and humorous text?\\ Give me 20 words or short phrases that show the literal style rather than \\ humorous style.\end{tabular}                                                                                          \\ \midrule
\multirow{3}{*}{Sarcasm} & sarcastic                                                 & \begin{tabular}[c]{@{}l@{}}Give me 10 words that show sarcastic style.\\ Give me 20 words or short phrases that people may use in text to show sarcasm.\end{tabular}                                                                                                                          \\
                         & literal                                                   & \begin{tabular}[c]{@{}l@{}}What's the difference between literal text and sarcastic text?\\ Give me 20 words or short phrases that show the literal style rather than sarcastic\\ style.\end{tabular}                                                                                         \\ \midrule
\multirow{26}{*}{Age}     & under12                                                   & \begin{tabular}[c]{@{}l@{}}Give me some words or phrases that an under-12-year-old child might say or write.\\ What words or phrases can a child under 12 say?\\ Imagine that you are 8 years old, what words or phrases do you often use in \\ communication and writing?\end{tabular}       \\
                         & 12-17                                                     & \begin{tabular}[c]{@{}l@{}}Give me some words or phrases that people aged 12-17 might say or write.\\ What words or phrases can a teenager aged 12-17 say?\\ Imagine that you are 14 years old, what words or phrases do you often use in\\ communication and writing?\end{tabular}           \\
                         & 18-24                                                     & \begin{tabular}[c]{@{}l@{}}Give me some words or phrases that people aged 18-24 might say or write.\\ What words or phrases can a person aged 18-24 say?\\ Imagine that you are 21 years old, what words or phrases do you often use in\\ communication and writing?\end{tabular}             \\
                         & 25-34                                                     & \begin{tabular}[c]{@{}l@{}}Give me some words or phrases that people aged 25-34 might say or write.\\ What words or phrases can a person aged 25-34 say?\\ Imagine that you are 30 years old, what words or phrases do you often use in\\ communication and writing?\end{tabular}             \\
                         & 35-44                                                     & \begin{tabular}[c]{@{}l@{}}Give me some words or phrases that people aged 35-44 might say or write.\\ What words or phrases can a person aged 35-44 say?\\ Imagine that you are 40 years old, what words or phrases do you often use in\\ communication and writing?\end{tabular}             \\
                         & 45-54                                                     & \begin{tabular}[c]{@{}l@{}}Give me some words or phrases that people aged 45-54 might say or write.\\ What words or phrases can a person aged 45-54 say?\\ Imagine that you are 50 years old, what words or phrases do you often use in\\ communication and writing?\end{tabular}             \\
                         & 55-74                                                     & \begin{tabular}[c]{@{}l@{}}Give me some words or phrases that people aged 55-74 might say or write.\\ What words or phrases can a person aged 55-74 say?\\ Imagine that you are 65 years old, what words or phrases do you often use in\\ communication and writing?\end{tabular}             \\
                         & \begin{tabular}[c]{@{}l@{}}75Years\\ OrOlder\end{tabular} & \begin{tabular}[c]{@{}l@{}}Give me some words or phrases that people aged 75 or older might say or write.\\ What words or phrases can a person aged 75 or older say?\\ Imagine that you are 80 years old, what words or phrases do you often use in\\ communication and writing?\end{tabular} \\ \bottomrule
\end{tabular}
\caption{Prompts used for ChatGPT to generate lexicon. Since we map the \emph{Age} dataset to a binary one, we also map the corresponding lexicons of its original age classes to the new classes. For example, the \emph{youthful} lexicon contains the contents of \emph{Under12}, \emph{12-17} and \emph{18-24} lexicons.
}
\label{tabs:chatgpt_prompt}
\end{table*}

\subsection{Statistics and Examples of Lexicons}
\label{appendix:lexicon_words_examples}
Table \ref{tab:lexicon_statistics} provides the statistics of NLP and ChatGPT lexicons used in the experiments.
Table \ref{tab:lexicon_examples} shows examples of lexicons from different sources. 
\begin{table}[t]
\begin{adjustbox}{width=\columnwidth}
\begin{tabular}{@{}lllc@{}}
\toprule
\multicolumn{1}{l}{\textbf{Style}} & \multicolumn{1}{l}{\textbf{Class}} & \multicolumn{1}{l}{\textbf{Lex Src}} & \textbf{\begin{tabular}[c]{@{}c@{}}Lex Size \\ (\# of words/phrases)\end{tabular}} \\ \midrule
Age                                & youthful                           & ChatGPT                              & 98                                                                                 \\
                                   & mature                             & ChatGPT                              & 65                                                                                 \\
Country                            & U.K                                & ChatGPT                              & 131                                                                                \\
                                   & U.S.A                              & ChatGPT                              & 127                                                                                \\
Formality                          & formal                             & NLP                                  & 330                                                                                \\
                                   & informal                           & NLP                                  & 370                                                                                \\
                                   & hate                               & NLP                                  & 178                                                                                \\
Hate/Offense                        & offensive                          & NLP                                  & 1403                                                                               \\
                                   & neither                            & ChatGPT                              & 5                                                                                  \\
Humor                              & humorous                           & ChatGPT                              & 21                                                                                 \\
                                   & literal                            & ChatGPT                              & 6                                                                                  \\
Politeness                         & polite                             & NLP                                  & 110                                                                                \\
                                   & impolite                           & ChatGPT                              & 54                                                                                 \\
Politics                           & LeftWing                           & NLP                                  & 2581                                                                               \\
                                   & Centrist                           & NLP                                  & 1231                                                                               \\
                                   & RightWing                          & NLP                                  & 2416                                                                               \\
Readability                        & simple                             & NLP                                  & 10290                                                                              \\
                                   & complex                            & NLP                                  & 4890                                                                               \\
Romance                            & romantic                           & ChatGPT                              & 58                                                                                 \\
                                   & literal                            & ChatGPT                              & 5                                                                                  \\
Sarcasm                            & sarcastic                          & ChatGPT                              & 34                                                                                 \\
                                   & literal                            & ChatGPT                              & 2                                                                                  \\
Sentiment                          & positive                           & NLP                                  & 204                                                                                \\
                                   & negative                           & NLP                                  & 292                                                                                \\
Shakespeare                        & shakespearean                      & NLP                                  & 1524                                                                               \\
                                   & modern                             & NLP                                  & 1524                                                                               \\
Subjectivity                       & subjective                         & NLP                                  & 5569                                                                               \\
                                   & objective                          & NLP                                  & 2653                                                                               \\ \bottomrule
\end{tabular}
\end{adjustbox}
\caption{Statistics of benchmark style lexicons.}
\label{tab:lexicon_statistics}
\end{table}
\begin{table}[t]
\centering
\begin{adjustbox}{width=\columnwidth}
\begin{tabular}{ll|llll}
\toprule
\textbf{Style}        & \textbf{Class}      & \textbf{Lex Src}                & \textbf{Lex}       \\ \midrule
                            &                            & \cellcolor[HTML]{FFFFFF}NLP  & \cellcolor[HTML]{FFFFFF}admittedly, albeit, insofar...                                                                              \\
                            & \multirow{-2}{*}{formal}   & \cellcolor[HTML]{EFEFEF}Dict    & \cellcolor[HTML]{EFEFEF}\begin{tabular}[c]{@{}l@{}}in accordance with rules of \\ convention or etiquette; official\end{tabular} \\ \cmidrule(l){2-4}  
                            &                            & \cellcolor[HTML]{FFFFFF}NLP  & \cellcolor[HTML]{FFFFFF}dude, kinda, sorta, repo...                  \\
\multirow{-5}{*}{Formality} & \multirow{-2}{*}{informal} & \cellcolor[HTML]{EFEFEF}Dict    & \cellcolor[HTML]{EFEFEF}\begin{tabular}[c]{@{}l@{}}having a relaxed, friendly, or \\ unofficial style\end{tabular}                  \\ \midrule
                            &                            & \cellcolor[HTML]{FFFFFF}ChatGPT & \cellcolor[HTML]{FFFFFF}funny, laugh-out-loud, silly...                                                                             \\
                            & \multirow{-1}{*}{humorous} & \cellcolor[HTML]{EFEFEF}Dict    & \cellcolor[HTML]{EFEFEF}being comical, amusing, witty                                                                               \\ 
                            &  & human    & chuckle, wisecrack, hilarious...
                                             \\ 
                            \cmidrule(l){2-4} 
                            &                            & \cellcolor[HTML]{FFFFFF}ChatGPT & \cellcolor[HTML]{FFFFFF}grim, formal, solemn, dour...                                                                               \\
\multirow{-4}{*}{Humor}     & \multirow{-2}{*}{literal}  & \cellcolor[HTML]{EFEFEF}Dict    & \cellcolor[HTML]{EFEFEF}not humorous; serious                                                                                       \\ &  & human    & analysis, scrutinize, enforce...\\\bottomrule
\end{tabular}
\end{adjustbox}
\caption{Examples of lexicons. ``Class" represents the category in a style. Each lexicon contains words or phrases that express or describe the class. ``Lex Src" indicates how the lexicon is collected (\S\ref{lexicon_collection}).}
\label{tab:lexicon_examples}
\end{table}

\section{Model-Generated Data For Generalization Experiments}
\label{appendix:novel_style_set_details}
Recall in \S\ref{sec:gpt4-styles} that in order to further evaluate the generalization capabilities of our porposed approach, we collected a diverse collection of styles using LLMs. Here we provide more details throughout the data generation process, including style creation (\S\ref{appendix:style_creation}), lexicon generation (\S\ref{appendix:lexicon_creation}), and labeled example (i.e., instance) generation (\S\ref{appendix:example_generation}). 

\subsection{Style Creation}
\label{appendix:style_creation}
We initiated the process of style classification task generation based on the thirteen styles outlined in our benchmark (refer to Table \ref{tab:dataset_statistics}). We had one author write the style classification instruction for each of these thirteen styles. During the task generation process, we randomly selected eight in-context examples from our pool, including three seed tasks and five model-generated tasks. 
We employed LLaMA-2-Chat 70B for new task generation. The template used for prompting these new style classification tasks are detailed in Table \ref{tab:task_generation_template}.
To ensure the diversity of the generated style classification tasks, a new task is added to the pool only when its ROUGE-L similarity with any existing task is less than 0.6. This process resulted in a total of 58 model-generated tasks, which we divided into 43 training tasks and 15 evaluation tasks. In order to further enrich the diversity of the evaluation task split, we designed 5 additional style classification tasks and incorporated them into the evaluation task split. Overall, this data generation process produces a total of 43 training style classification tasks and 20 evaluation style classification tasks. We present the full list of 63 generated style classification tasks in Table \ref{appendix:full_list_63_styles}.

\subsection{Lexicon Creation}
\label{appendix:lexicon_creation}
After creating the training and evaluation tasks, we employed LLaMA-2-Chat 70B to generate a concise lexicon for each class in the style classification tasks, using in-context examples. Our ablation studies, as detailed in \S\ref{appendix:lexicon_size}, revealed that a lexicon consisting of just five words or phrases are sufficient for effective generalization to new styles. So we restricted the lexicon size for each style class to five words or phrases. The template used for prompting the generation of style class lexicons are displayed in Table \ref{tab:generalization_lexicon_template}.

\subsection{Labeled Example Generation}
\label{appendix:example_generation}
We prompted LLaMA-2-Chat 7B to generate labeled examples for our training style classification tasks, and GPT-4 to generate examples for our evaluation tasks. Both utilize the same prompting template presented in Table \ref{tab:example_generation_template} for labeled example generation.

\begin{table*}[t]
    \centering
    \small
    \noindent\fbox{%
    \begin{minipage}{\dimexpr\linewidth-2\fboxrule-2\fboxsep-100pt\relax} 
\tt 
Come up with a series of textual classification tasks about writing styles. Try to specify the possible output labels when possible.\\
\\
Task 1: \{instruction for existing task 1\} \\
Task 2: \{instruction for existing task 2\} \\
Task 3: \{instruction for existing task 3\} \\
Task 4: \{instruction for existing task 4\} \\
Task 5: \{instruction for existing task 5\} \\
Task 6: \{instruction for existing task 6\} \\
Task 7: \{instruction for existing task 7\} \\
Task 8: \{instruction for existing task 8\} \\
Task 9:
\end{minipage}}
\caption{Prompt template used for generating new style classification tasks. 8 existing instructions are randomly sampled from the task pool for in-context demonstration. The model is allowed to generate instructions for new tasks, until it stops its generation or reaches its length limit.}
\label{tab:task_generation_template}
\end{table*}

\begin{table*}[]
    \centering
    \small
    \noindent\fbox{%
    \begin{minipage}{\dimexpr\linewidth-2\fboxrule-2\fboxsep-100pt\relax} 
\tt 
You are a helpful AI assistant. Generate a few words that describe or exhibit the target style. If the words cannot fully express the characteristics of the style, define the style with phrases or short sentences. \\
\\
Example\\
Style class 1: \{lexicon words/phrases for style class 1\} \\
\\
Example\\
Style class 2: \{lexicon words/phrases for style class 2\} \\
\\
$\cdots$ \\
\\
Example\\
Style class 8: \{lexicon words/phrases for style class 8\} \\
\\
Example\\
Style class 9:
\end{minipage}}
\caption{Prompt template used for generating style class lexicon.}
\label{tab:generalization_lexicon_template}
\end{table*}

\begin{table*}[]
    \centering
    \small
    \noindent\fbox{%
    \begin{minipage}{\dimexpr\linewidth-2\fboxrule-2\fboxsep-15pt\relax}
\tt
You are a helpful AI assistant. Given the classification task definition and the possible output labels, generate an input that corresponds to each of the class labels. Try to generate high-quality inputs with varying lengths.\\
\\
{Task: Classify the sentiment of a sentence. The possible output labels are: positive, negative.} \\
Label: positive \\
Sentence: I had a great day today. The weather was beautiful and I spent time with friends and family.\\
Label: negative\\
Sentence: I was really disappointed by the latest superhero movie.\\
\\
Task: Categorize the writing style of a given piece of text into romantic, or not romantic.\\
Label: romantic\\
Text: A lot of people spend their whole lives looking for true love and ultimately fail. So how ungrateful would I be, if I let our love fade? That~@~Ys how you know, my love is here to stay.\\
Label: not romantic\\
Text: I need you to submit this proposal as soon as possible.\\
\\
$\cdots$ \\
\\
Task: \{instruction for the target task\}
\end{minipage}}
\caption{Prompt template used for generating the example for classification tasks.}
\label{tab:example_generation_template}
\end{table*}

\onecolumn
\begin{center}
\tablefirsthead{%
\hline
\footnotesize{\textbf{Style Classification Task}}& \footnotesize{\textbf{Classes}} \\
\hline}
\tablehead{%
\hline
\multicolumn{2}{l}{\small\sl continued from previous page}\\
\hline
 \footnotesize{\textbf{Style Classification Task}}& \footnotesize{\textbf{Classes}}\\
\hline}
\tabletail{%
\hline
\multicolumn{2}{r}{\small\sl continued on next page}\\
\hline}
\tablelasttail{\hline}
\bottomcaption{63 generated style classification tasks in \S\ref{sec:gpt4-styles}.}
\label{appendix:full_list_63_styles}
	\begin{supertabular}{ll}
		\hline
		\scriptsize {Identify the type of writing style used in a given text.} &\scriptsize {narrative, descriptive, expository, persuasive}   \\ \shrinkheight{10pt} \hline
			\scriptsize {Determine whether the given text contains any errors in grammar, spelling, or punctuation.} &\scriptsize {error-free, erroneous}       \\ \hline
			\scriptsize {Classify the style of a poem into one of the four types. } &\scriptsize {sonnet, haiku, free verse, limerick}       \\ \hline\scriptsize {Categorize the emotion of the utterances.} &\scriptsize {angry, disgusted, fearful, happy, sad}       \\ \hline\scriptsize {Determine the level of organization in the text.} &\scriptsize {well-organized, disorganized}       \\ \hline\scriptsize {Classify the style of a text according to its structure.} &\scriptsize {chronological, non-chronological}       \\ \hline\scriptsize {Classify the text according to its tone.} &\scriptsize {friendly, hostile, neutral}       \\ \hline
   \scriptsize {\begin{tabular}[c]{@{}l@{}}Define the writing style "Infotainment" as "merging informative writing with an entertaining \\ approach". Define the writing  style "Techeative" as "blending technical writing (e.g. precise \\ descriptions of complex subjects) with creative elements to  make it more engaging and \\ understandable". Classify  the style of a presentation into one of the above two categories.\end{tabular}} &\scriptsize {Infotainment, Techeative}       \\ \hline\scriptsize {Classify the style of a text according to its content and language use.} &\scriptsize {rational, irrational}       \\ \hline\scriptsize {Evaluate the level of clarity in the text.} &\scriptsize {clear, unclear}       \\ \hline\scriptsize {Classify the text style according to its tone and language use.} &\scriptsize {nostalgic, reflective, analytical}       \\ \hline\scriptsize {Classify the style of a text according to its content and language use.} &\scriptsize {creative, conventional}       \\ \hline\scriptsize {Identify the author's voice style in a given text.} &\scriptsize {authoritative, unreliable}       \\ \hline\scriptsize {Evaluate the level of emotional appeal in the text.} &\scriptsize {low emotional appeal, high emotional appeal}       \\ \hline\scriptsize {Determine the level of originality in a story.} &\scriptsize {original, somewhat original, not original}       \\ \hline\scriptsize {Evaluate the level of credibility in the text.} &\scriptsize {credible, moderately credible, not credible}       \\ \hline\scriptsize {Read a passage, and select the topic for this passage based on the content and text style.} &\scriptsize {\begin{tabular}[c]{@{}l@{}}finance, politics, health, education, \\ technology, entertainment\end{tabular}}       \\ \hline\scriptsize {Read the summary of a book and categorize its genre.} &\scriptsize {science fiction, romance, thriller, biography}       \\ \hline\scriptsize {Determine the primary intention behind the author's writing of a specific text.} &\scriptsize {persuasive, informative, entertaining, educational}       \\ \hline\scriptsize {Classify the text style according to its tone and language use.} &\scriptsize {assertive, submissive}       \\ \hline\scriptsize {Classify the text style according to its tone and language use.} &\scriptsize {strong, weak}       \\ \hline\scriptsize {Classify text style.} &\scriptsize {conversational, academic}       \\ \hline\scriptsize {Determine the most likely author based on the writing style.} &\scriptsize {Hemingway, Joyce, Kafka, Hurston, Christie}       \\ \hline\scriptsize {Classify the text style according to its tone and language use.} &\scriptsize {monotonous, engaging}       \\ \hline\scriptsize {Classify the content of a piece of text.} &\scriptsize {spam, ham}       \\ \hline\scriptsize {Read the text and classify its style.} &\scriptsize {fictional, non-fictional}       \\ \hline\scriptsize {Evaluate the mood of a song based on its lyrics.} &\scriptsize {relaxing, energizing, romantic, melancholic}       \\ \hline\scriptsize {Identify the rhetorical devices used in a given text.} &\scriptsize {\begin{tabular}[c]{@{}l@{}}onomatopoeia, alliteration, hyperbole,\\ repetition, oxymoron\end{tabular} }       \\ \hline\scriptsize {Classify the text as one of the following: journalistic, academic, or literary.} &\scriptsize {journalistic, academic, literary}       \\ \hline\scriptsize {Assess how supportive the context is in response to a request for help.} &\scriptsize {very supportive, moderately supportive, not supportive}       \\ \hline\scriptsize {Classify the text according to its tone and language use.} &\scriptsize {realistic, idealistic}       \\ \hline\scriptsize {Given a famous quote, classify its tone style into one of the four categories.} &\scriptsize {inspirational, funny, philosophical, sarcastic}       \\ \hline\scriptsize {Classify the text according to its tone and language use.} &\scriptsize {confident, uncertain, timid}       \\ \hline\scriptsize {Carefully review the provided text and assess its level of rigor.} &\scriptsize {rigorous, careless}       \\ \hline\scriptsize {Classify the author's attitude towards the topic.} &\scriptsize {enthusiastic, uninterested}       \\ \hline\scriptsize {Assess the difficulty level of academic texts, and choose the label from the following four options. } &\scriptsize {elementary, intermediate, advanced, expert}       \\ \hline\scriptsize {Analyze the given text and determine whether it contains any biases.} &\scriptsize {biased, unbiased}       \\\hline\scriptsize {Classify the style of an example.} &\scriptsize {adventurous, cautious, conservative}       \\ \hline\scriptsize {Classify the text according to its tone.} &\scriptsize {optimistic, pessimistic, neutral}       \\ \hline\scriptsize {Classify the text style.} &\scriptsize {logical, emotional}       \\ \hline\scriptsize {Classify the text according to its tone and language style.} &\scriptsize {apologetic, accusatory, grateful, condescending}       \\ \hline\scriptsize {Determine the response style by examining the content and the quality of a response.} &\scriptsize {\begin{tabular}[c]{@{}l@{}}helpful and harmless, helpful and harmful,\\ helpless and harmless, helpless and harmful\end{tabular}}       \\ \hline\scriptsize {\begin{tabular}[c]{@{}l@{}}Identify the style of a sonnet by analyzing the rhyme scheme of its first four lines, each separated \\ by a newline symbol.\end{tabular}} &\scriptsize {Shakespearean sonnet, Petrarchan sonnet.}       \\ \hline\scriptsize {\begin{tabular}[c]{@{}l@{}}Identify the style of a poetry by analyzing the rhyme scheme of its first four lines, each separated\\ by a newline symbol.\end{tabular}} &\scriptsize {ABAB, AABB}       \\ \hline\scriptsize {Carefully review the provided text and determine the nature of its writing style.} &\scriptsize {machine-generated text, human-written text}       \\ \hline\scriptsize {\begin{tabular}[c]{@{}l@{}}XXX and YYY are two Ph.D. students who often engage in writing papers. XXX has a penchant \\ for employing a variety of fancy words and clauses in the writing, whereas YYY favors a style \\ that is more concise and straightforward, focusing on brevity and clarity. Given a piece of text,\\ determine who is  more likely to be the author based on the writing style.\end{tabular} } &\scriptsize {XXX, YYY}       \\ \hline\scriptsize {Determine the level of coherence in a piece of writing.} &\scriptsize {coherent, incoherent}       \\ \hline\scriptsize {\begin{tabular}[c]{@{}l@{}}Determine whether the text contains any sensitive information such as personal data, financial \\ information, or explicit content.\end{tabular}} &\scriptsize {sensitive, non-sensitive}       \\ \hline\scriptsize {Classify text format based on the language style used.} &\scriptsize {editorial, blog post, research paper, poem, script}       \\ \hline\scriptsize {Determine if a tweet contains misinformation.} &\scriptsize {true, misleading}       \\ \hline\scriptsize {Determine the level of nuance in a piece of writing.} &\scriptsize {nuanced, somewhat nuanced, not nuanced}       \\ \hline\scriptsize {Classify text style according to its intended audience.} &\scriptsize {general public, experts, children, young adults}       \\ \hline\scriptsize {Analyze the tone of a customer review for a product.} &\scriptsize {satisfied, dissatisfied, mixed feelings}       \\ \hline\scriptsize {Determine the tone of the text. } &\scriptsize {serious, ironic, condescending}       \\ \hline\scriptsize {Evaluate the level of technical jargon used in the text.} &\scriptsize {technical, non-technical}       \\ \hline\scriptsize {Classify the attitude of the author into either \"wanting to help\" or \"perfunctory\".} &\scriptsize {helpful, unhelpful}       \\ \hline\scriptsize {Classify the poetry style type.} &\scriptsize {ballad, acrostic, ode, elegy, limerick}       \\ \hline\scriptsize {\begin{tabular}[c]{@{}l@{}}Define the style of a "empathetic, colloquial, humorous, lively" response as "teddy bear". \\ Define the style of a "calm, caring, professional, earnest" response as "psyduck". \\ Classify the style of responses made by a senior AI Assistant.\end{tabular}} &\scriptsize {teddy bear, psyduck}       \\ \hline\scriptsize {Analyze the content and language style of the support ticket or email and classify its urgency level.} &\scriptsize {\begin{tabular}[c]{@{}l@{}}high urgency, medium urgency, low urgency, \\ informational\end{tabular}}       \\ \hline\scriptsize {Given a sentence, detect if there is any potential stereotype in it.} &\scriptsize {stereotyped, non-stereotyped}       \\ \hline\scriptsize {Determine the level of conciseness in a piece of writing.} &\scriptsize {concise, verbose}       \\ \hline\scriptsize {\begin{tabular}[c]{@{}l@{}}A desirable trait in a human-facing dialogue agent is to appropriately respond to a conversation \\ partner that is describing personal experiences, by understanding and acknowledging any \\ implied feelings - a skill we refer to as empathetic responding. Classify the response style.\end{tabular}} &\scriptsize {empathetic, indifferent}       \\ \hline\scriptsize {Identify the rhetorical devices used in a given text.} &\scriptsize {metaphor, simile, personification}       \\ \hline
\end{supertabular}
\end{center}
\twocolumn

\subsection{Data Quality}
\label{appendix:quality_check_eval_set}
To investigate the quality of our generated evaluation set $\mathcal{D}_{\text{eval}}$, we randomly sampled a shared set of 500 annotation examples. We asked each annotator to assess the quality of the examples and their labels. Specifically, they were instructed to consider these questions: (1) is it easy to identify a style (e.g., low emotional appeal vs. high emotional appeal) in a given text example? (2) is the label of the given example accurate? (3) If the label is incorrect, what would be the correct label?

We collected responses from three annotators. According to their quality reviews, about 97\% of the examples are easily identifiable by their styles, with all annotators answering ``Yes" to the first question. Gathering responses for the last two questions from three annotators provided three human annotations for each example. We followed prior works \cite{kang2021xslue, artstein2008inter} to compute the inter-annotator agreement using Krippendorff’s alpha \cite{krippendorff2004reliability} for these annotation examples. A high agreement score of 93.27\% indicates strong reliability in the annotations \cite{krippendorff2004reliability}. Moreover, our further analysis reveals that 87.4\% of the evaluated examples received unanimous agreement the GPT-4 generated label was correct from all three annotators, and 99.2\% reached a consensus among at least two annotators,  highlighting a high level of agreement and consistency in the annotation process.

\subsection{Statistics and Examples of Generated Data}
\label{appendix:eval_set_examples}
\begin{figure}[]
    \centering
    \includegraphics[width=\columnwidth]{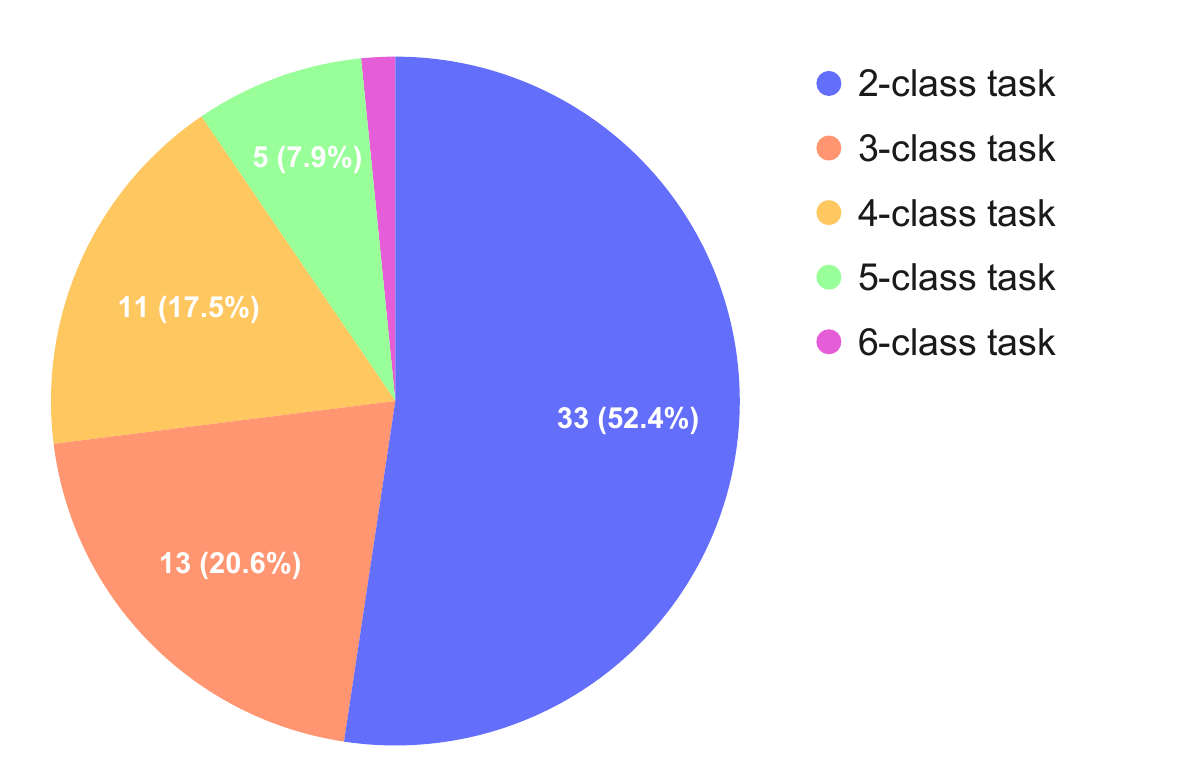}
    \caption{Distribution of 63 style classification tasks in \S\ref{sec:gpt4-styles}.}
    \label{fig:case_study_task_distribution}
\end{figure}

Figure \ref{fig:case_study_task_distribution} plots the distribution of 63 generated style classification tasks in this data generation process. We present examples of style annotation data and their lexicons in Table \ref{tab:case_study_set_example}.
\begin{table*}[]
\begin{adjustbox}{width=2\columnwidth}
\centering
\begin{tabular}{lll}
\hline
\toprule
\textbf{Style Classes and their Lexicons}                                                                                                                                                                                                                                               & \textbf{Example}                                                                                                                                                                                                                             & \textbf{Label} \\ \midrule
\multirow{2}{*}{\begin{tabular}[c]{@{}l@{}}\textbf{helpful}: supportive, wanting to help\\ \textbf{unhelpful}: perfunctory, unfavorable\end{tabular}}                                                                                                                                                                 & Okay, save it. I don't have time to hear your complaints.                                                                                                                                                                                    & unhelpful      \\ \cmidrule(l){2-3} 
                                                                                                                                                                                                                                                                                        & \begin{tabular}[c]{@{}l@{}}Person A: "I've been having a hard time getting over my ex."\\ Person B: "Healing takes time, and it's okay to grieve a relationship. If you need \\ someone to talk to, I'm here for you, anytime."\end{tabular} & helpful        \\ \midrule
\multirow{3}{*}{\begin{tabular}[c]{@{}l@{}}\textbf{acrostic}: nitials, word puzzle, creative\\ \textbf{ghazal}: lyrical, emotive, spiritual\\ \textbf{limerick}: humorous, rhythmic, short\end{tabular}}                                                                                                         & \begin{tabular}[c]{@{}l@{}}There once was a man from Nantucket\\ Who kept all his cash in a bucket.\\ But his daughter, named Nan,\\ Ran away with a man\\ And as for the bucket, Nantucket.\end{tabular}                                    & limerick       \\ \cmidrule(l){2-3} 
                                                                                                                                                                                                                                                                                        & \begin{tabular}[c]{@{}l@{}}I am lost in love's reality, and I see you in dreams,\\ In the silence of the night, in the roar of the streams, it's you.\end{tabular}                                                                           & ghazal         \\ \cmidrule(l){2-3} 
                                                                                                                                                                                                                                                                                        & \begin{tabular}[c]{@{}l@{}}Caring and kind,\\ Always in my mind.\\ Today and tomorrow,\\ Heart full of sorrow.\\ Yearning for your touch.\end{tabular}                                                                                       & acrostic       \\ \midrule
\multirow{2}{*}{\begin{tabular}[c]{@{}l@{}}\textbf{supportive}: empathetic, encouraging, comforting, helpful\\ \textbf{unsupportive}: distant, dismissive, uncaring, brief\end{tabular}}                                                                                                                  & I believe in your abilities and I know you can do it.                                                                                                                                                                                        & supportive     \\ \cmidrule(l){2-3} 
                                                                                                                                                                                                                                                                                        & That's not up to the mark. You need to work harder.                                                                                                                                                                                          & unsupportive   \\ \midrule
\multirow{3}{*}{\begin{tabular}[c]{@{}l@{}}\textbf{philosophical}: relating to the fundamental nature of \\                        knowledge, reality, and existence\\ \textbf{inspirational}: providing creative or spiritual inspiration\\ \textbf{funny}: humorous, causing laughter or amusement\end{tabular}} & It does not matter how slowly you go as long as you do not stop.                                                                                                                                                                             & inspirational  \\ \cmidrule(l){2-3} 
                                                                                                                                                                                                                                                                                        & The unexamined life is not worth living.                                                                                                                                                                                                     & philosophical  \\ \cmidrule(l){2-3} 
                                                                                                                                                                                                                                                                                        & \begin{tabular}[c]{@{}l@{}}I find television very educating. Every time somebody turns on the set, I go into\\ the other room and read a book.\end{tabular}                                                                                  & funny          \\ \midrule
\multirow{2}{*}{\begin{tabular}[c]{@{}l@{}}\textbf{condescending}: patronizing, arrogant, superior\\ \textbf{respectful}: polite, considerate, humble\end{tabular}}                                                                                                                                       & Wow, you actually understood that concept? I'm impressed.                                                                                                                                                                                    & condescending  \\ \cmidrule(l){2-3} 
                                                                                                                                                                                                                                                                                        & \begin{tabular}[c]{@{}l@{}}Your social life seems vibrant and you're also doing well in your work. How do \\ you manage that?\end{tabular}                                                                                                   & respectful     \\ \bottomrule
\end{tabular}
\end{adjustbox}
\caption{Examples of new styles and instances generated semi-automatically using LLMs.  These styles are used in \S \ref{sec:gpt4-styles} to further demonstrate the generalization ability of lexicon-based instructions.}
\label{tab:case_study_set_example}
\end{table*}

\section{Implementation Details}
\label{appendix:implementation_details}
We use PyTorch \cite{pytorch} and Huggingface Transformers \cite{wolf-etal-2020-transformers} in the experiments.

In our preliminary studies, we compare the performance of fully fine-tuned, partially (only the last two layers) fine-tuned, and parameter-efficiently fine-tuned GPT-J. We find that fine-tuning GPT-J with LoRA \cite{hu2021lora} achieves the best performance, so we use this method in our main experiments with GPT-J. 

In the zero-shot learning experiments, we prompted LLaMA-2-Chat (13B) to predict the target styles without any fine-tuning. We employed 4-bit inference due to our computing resource constraints \cite{dettmers2023case}.

In the zero-shot cross-style experiments, we first fine-tuned a model on the training styles before evaluating it on the evaluation styles. We fine-tuned the LLaMA-2 (7B) model on 4 A40 GPUs using DeepSpeed. All the other models were fine-tuned on one single A40 GPU. 
Hyperparameters are selected following the common practices in previous research.
Table \ref{hyperparameters} reports the hyperparameters for our instruction tuning.


\begin{table*}[t]
\centering
\small
\begin{tabular}{llll}
\toprule
\textbf{Hyperparameter}   & \textbf{T5$_\text{base}$}                                          & \textbf{GPT-J} &\textbf{LLaMA-2 7B} \\ \midrule
optimizer                 & Adafactor                                                         & Adam      & Adam    \\
learning rate             & 1e-4                                                              & 1e-5       & 2e-5    \\
batch size                & 8                                                                & 4          &  128  \\
max encoder/input length  & 512                                                          & 512         & 512   \\
max decoder/target length & 16                                                                & 16         &    \\
\# epochs                 & \begin{tabular}[c]{@{}l@{}}Instruction with class randomization: 5\\ Others: 3\end{tabular} & 1     & 3         \\ \bottomrule
\end{tabular}
\caption{Hyperparameters of instruction tuning on the benchmark training styles. Note that the number of epochs depends on the model convergence rate. Instruction with class name randomization converge more slowly than the other prompts, so their epoch is longer.}
\label{hyperparameters}
\end{table*}


\section{Additional Experimental Results \& Analyses}
\label{appendix:more_zeroshot_results}
\begin{table}[ht]
\setlength{\tabcolsep}{2pt}
\centering
\begin{adjustbox}{width=\linewidth}
\begin{tabular}{llcccccc}
\toprule
   &   & \texttt{\sOne}   & \texttt{\sTwo}  & \texttt{\sThree} & \texttt{\sFour}  & \cellcolor[HTML]{D9D9D9}\textbf{Avg.} & \cellcolor[HTML]{D9D9D9}\textbf{SD.} \\ \midrule
\multicolumn{1}{l}{\multirow{6}{*}{\begin{tabular}[c]{@{}l@{}}w/o\\ rand.\\({\small\emph{{Lang}}})\end{tabular}}}                      & \texttt{t1}                                    & 42.55                         & 43.88                       & 41.99          &  41.64          & \cellcolor[HTML]{D9D9D9} 42.52        & \cellcolor[HTML]{D9D9D9}0.99         \\
\multicolumn{1}{l}{}                      & \texttt{t2}                                    & 39.05                         & 41.72                        & 33.71        & 41.56                & \cellcolor[HTML]{D9D9D9}39.01        & \cellcolor[HTML]{D9D9D9}3.74         \\
\multicolumn{1}{l}{}                      & \texttt{t3}                                   & 35.40                         & 40.21                         & 36.13     &38.69                  & \cellcolor[HTML]{D9D9D9}37.61        & \cellcolor[HTML]{D9D9D9}2.24         \\
\multicolumn{1}{l}{}                      & \texttt{t4}                                   & 30.43                        & 36.33                     & 37.02    &36.14                   & \cellcolor[HTML]{D9D9D9}34.98       & \cellcolor[HTML]{D9D9D9}3.06         \\
\multicolumn{1}{l}{}                      & \cellcolor[HTML]{C0C0C0}\textbf{Avg.} & \cellcolor[HTML]{C0C0C0}36.86 & \cellcolor[HTML]{C0C0C0}40.54 & \cellcolor[HTML]{C0C0C0}37.21 & \cellcolor[HTML]{C0C0C0}39.51 &                                     &                                      \\
\multicolumn{1}{l}{} & \cellcolor[HTML]{C0C0C0}\textbf{SD.}  & \cellcolor[HTML]{C0C0C0}5.18  & \cellcolor[HTML]{C0C0C0}3.18 & \cellcolor[HTML]{C0C0C0}3.48  & \cellcolor[HTML]{C0C0C0}2.63 &                                       &                                      \\ \midrule
\multicolumn{1}{l}{\multirow{6}{*}{\begin{tabular}[c]{@{}l@{}}w/\\ rand.\\({\small\emph{{Lang, Rw}}})\end{tabular}}}                      & \texttt{t1}                                 & 54.20                         & 54.72                      & 53.16  & 55.15                     & \cellcolor[HTML]{D9D9D9}54.31        & \cellcolor[HTML]{D9D9D9}0.86         \\
\multicolumn{1}{l}{}                      &\texttt{t2}                                 & 54.74                        & 54.23                         & 50.24    &54.83                    & \cellcolor[HTML]{D9D9D9}53.51        & \cellcolor[HTML]{D9D9D9}2.20        \\
\multicolumn{1}{l}{}                      &\texttt{t3}                               & 53.24                         & 52.17                        & 51.59     &53.85                    & \cellcolor[HTML]{D9D9D9}52.71        & \cellcolor[HTML]{D9D9D9}1.02       \\
\multicolumn{1}{l}{}                      &\texttt{t4}                        & 55.83                         & 51.89                         & 55.02        &53.91                 & \cellcolor[HTML]{D9D9D9}54.16         & \cellcolor[HTML]{D9D9D9}1.71         \\
\multicolumn{1}{l}{}                      & \cellcolor[HTML]{C0C0C0}\textbf{Avg.} & \cellcolor[HTML]{C0C0C0}54.50 & \cellcolor[HTML]{C0C0C0}53.25 & \cellcolor[HTML]{C0C0C0}52.50 &\cellcolor[HTML]{C0C0C0}54.44 &                                       &                                      \\
\multicolumn{1}{l}{} & \cellcolor[HTML]{C0C0C0}\textbf{SD.}  & \cellcolor[HTML]{C0C0C0}1.08  & \cellcolor[HTML]{C0C0C0}1.43  & \cellcolor[HTML]{C0C0C0}2.06  &\cellcolor[HTML]{C0C0C0}0.66 &               &                                      \\ \bottomrule
\end{tabular}
\end{adjustbox}
\caption{For each combination of the lexicon source and the prompt template, class randomization (i.e., the ``Lang, Rw'' variant) consistently improves the average F1 scores. \texttt{t1}, \texttt{t2}, \texttt{t3} and \texttt{t4} are the different templates detailed in Table \ref{tab:prompts_examples}. \texttt{{\sOne}, \texttt{\sTwo}, \texttt{\sThree} and \texttt{\sFour}} are the different lexicon sources described in Appendix \ref{ablate:lex_source}. Each white cell reports the result averaged over the six target styles. \lightGrey{Light grey} cells indicate the average (Avg.) and the standard deviation (SD.) scores over four lexicon sources. \darkGrey{Dark grey} cells represent Avg. and SD. over four templates. }
\label{tab:ablation_prompt_templates}
\end{table}
\subsection{Impact of Instruction Templates}
\label{prompt_sensitivity}
Prior works find that prompting an LLM on an unseen task is extremely sensitive to the prompt design, such as the wording of prompts \cite{t0}. To investigate the sensitivity of lexicon-based instructions, we experiment with four instruction templates \texttt{t1}, \texttt{t2}, \texttt{t3}, \texttt{t4} (Table \ref{tab:prompts_examples}), each of which contains different natural language task instructions. For each template, we fine-tune a model on our benchmark training styles using lexicon-based instructions. Table \ref{tab:ablation_prompt_templates} shows that without randomization during instruction tuning, lexicon-based instruction (i.e., the ``Lang'' variant) is sensitive to the choice of templates. However, after introducing
class randomization, lexicon-based instruction (i.e., the ``Lang, Rw'' variant) improves the average F1 across the templates by a substantial margin, while reducing the standard deviation, indicating that it is more robust to the wordings of the prompts.

\paragraph{Instruction Template in Main Experiments}
\label{appendix:main_exp_template}
In our main experiments (\S\ref{experimental_results}), we conduct a comparative analysis between the lexicon-based instruction and the standard instruction. Both utilize the template \textbf{t2} in Table \ref{tab:prompts_examples} except that the standard instruction does not incorporate any lexicon sampling. Instead, each slot for the lexicon words contains only the corresponding class name. Here is an example input of the standard instruction on Politeness: \emph{In this task, you are given sentences. The task is to classify a sentence as "polite" if the style of the sentence is similar to the words "polite" or as "impolite" if the style of the sentence is similar to the words "impolite". Here is the sentence: "I've just noticed I wrote... and smooth out the text?"}. Its output is \emph{polite}.

\begin{figure*}[t]
\centering
    \includegraphics[width=\textwidth]{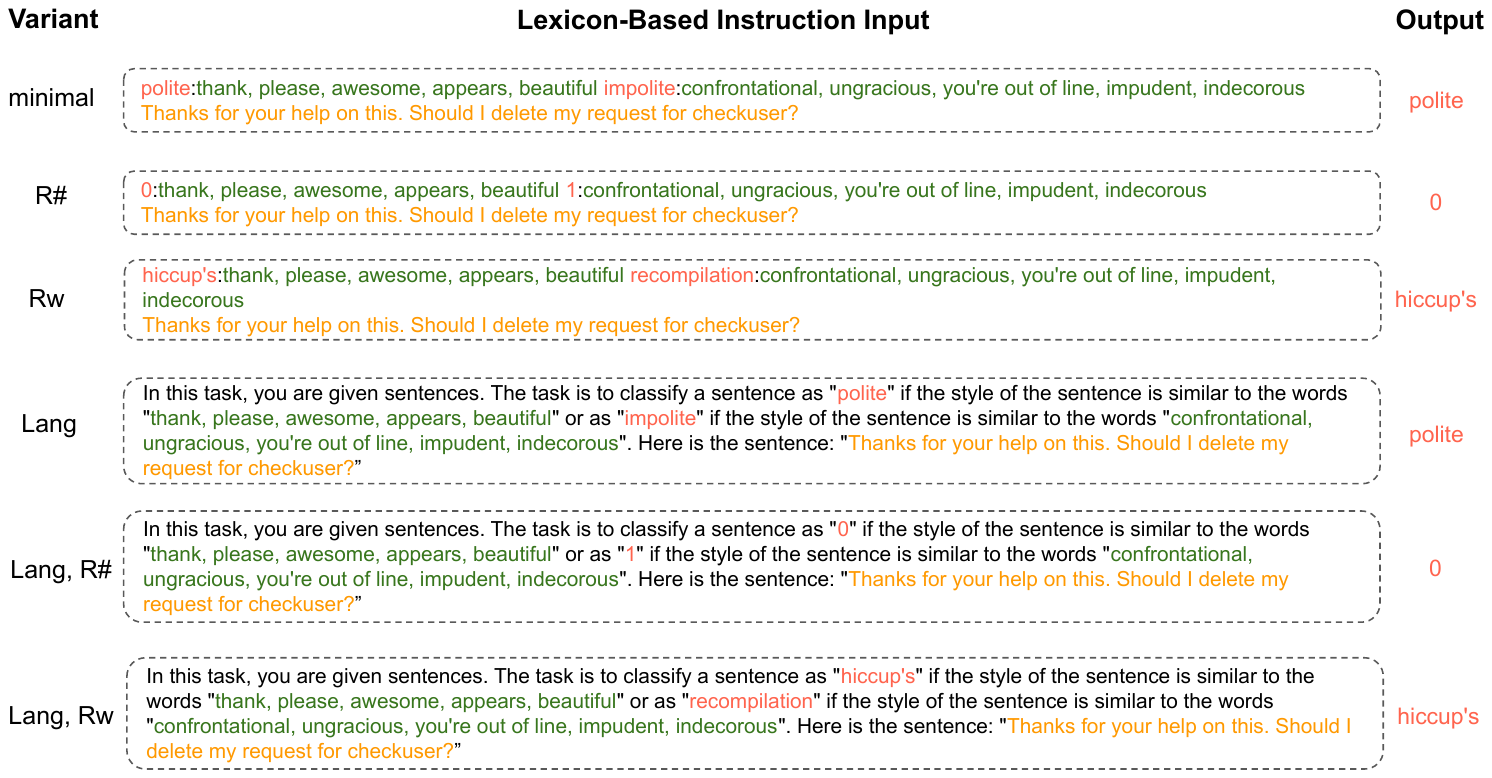}
    \vskip -0.8in
    \caption{Examples of different lexicon-based instruction variants (as detailed in \S\ref{variant_lps}) on \emph{Politeness}. \textcolor{randClassForMethod2}{Red} part is (randomized) classes, the \textcolor{subsetLexForMethod}{green} part represents the words sampled from each class lexicon, and \textcolor{inputSentForMethod}{yellow} stands for the input sentence and the uncolored part is the instruction template. 
    }
    \label{fig:lp_examples}
\end{figure*}
\begin{figure}[t]
    \centering
    \includegraphics[width=0.85\columnwidth]{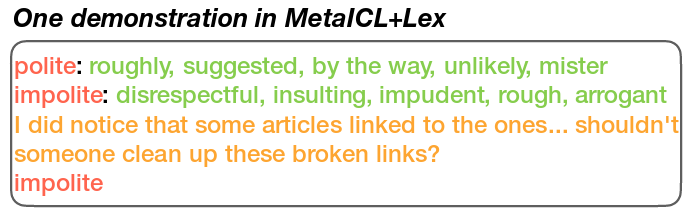}
    \caption{MetaICL+Lex input consists of $K$ demonstrations and an input sentence. Each demonstration contains $m$ \textcolor{subsetLexForMethod}{lexicon words} for each \textcolor{randClassForMethod2}{class}, followed by an \textcolor{inputSentForMethod}{example} with its \textcolor{randClassForMethod2}{label}.
    }
    \label{fig:metaicl+lex_example.png}
\end{figure} 

\subsection{Impact of Lexicon Source}
\label{ablate:lex_source}
We study the impact of lexicon choices in lexicon-based instruction that include: (1) \texttt{\sOne}: all lexicons are from dictionary; (2) \texttt{\sTwo}: for classes that have NLP lexicons, we directly use them, whereas for those without, we create ones using ChatGPT; (3) \texttt{\sThree}: each class lexicon contains only its class name, e.g., the \say{humorous} lexicon has a single word \say{humorous}; (4) \texttt{\sFour}: we have a native speaker create a lexicon for each style class, by carefully choosing words or phrases that best capture the characteristics of each style class.
Table \ref{tab:ablation_prompt_templates} shows that without class randomization during instruction tuning with lexicon, the average F1 for \texttt{\sTwo} across four templates is the highest at 40.54. With randomization, \texttt{\sOne} performs the best at 54.50. Randomizing classes with words in lexicon-based instructions consistently improves the average F1 while reducing the standard deviation across four lexicon sources, regardless of the prompt templates used. The human-created lexicon is the most robust to the change of templates.

\subsection{Varying Number of Lexicon Words (m) in Lexicon-Based Instructions}
\label{appendix:lexicon_size}
\begin{figure}[t]
    \centering
    \includegraphics[width=\columnwidth]{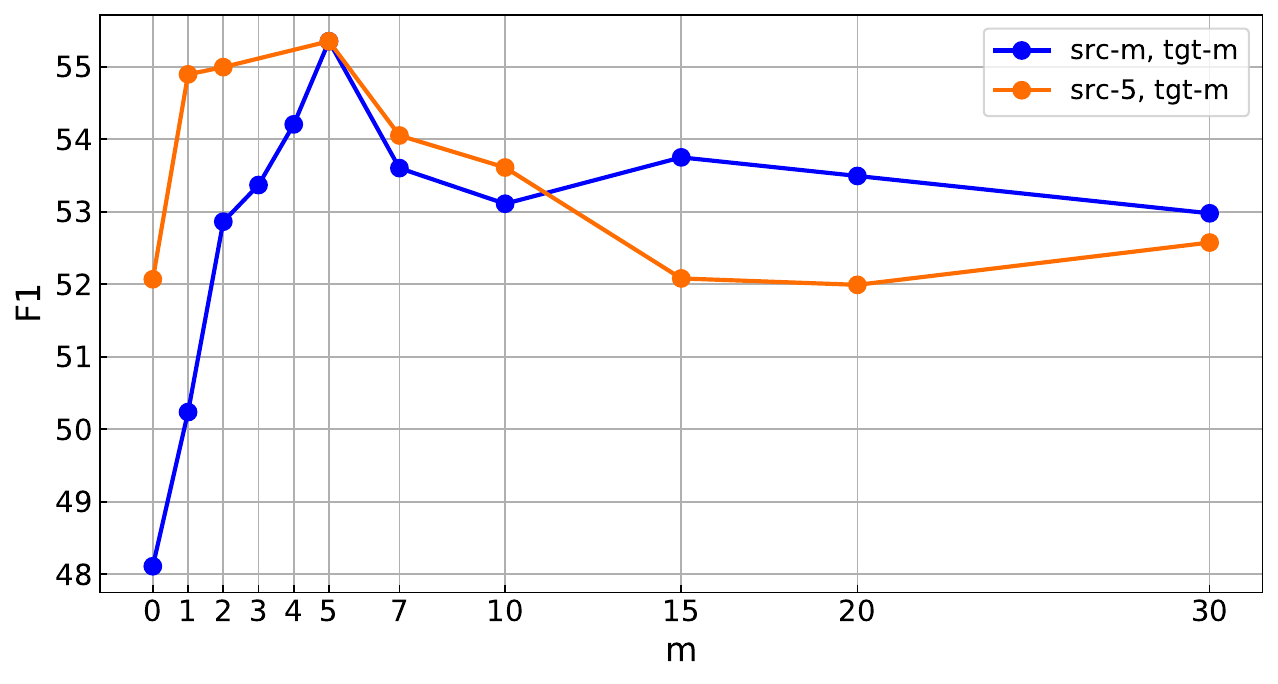}
    \caption{
    Impact of the number ($m$) of lexicon words or phrases used in each lexicon-based instruction. ``src-$m$" is for fine-tuning on source styles (i.e., training styles) and ``tgt-$m$" for evaluation on targets.}
    \label{fig:ablation_lex_size_train_test}
\end{figure}

When predicting a style in the evaluation split zero-shot, the  lexicon instruction-tuned model only has access to a subset of $m$ lexicon words that express or imply the style classes rather than example sentences. 
To investigate the model's dependence on the number of lexicon words, we take the variant of lexicon-based instruction with class randomization (i.e., the ``Lang, Rw'' variant) and incrementally increase $m$ from $0$ to $30$ in both fine-tuning and evaluation phases. Figure \ref{fig:ablation_lex_size_train_test} shows a general trend that the average F1 of six targets initially increases with increasing $m$, but then either drops or stabilizes.  
On average, our method performs the best when $m=5$.

Moreover, we fix the model fine-tuned with the ``Lang, Rw" lexicon-based instruction variant at $m=5$, and then gradually increase $m$ while evaluating evaluation styles.  A similar trend is noticed in Figure \ref{fig:ablation_lex_size_train_test}. It can also be seen that when target styles have no lexicon resources ($m=0$), increasing the number of lexicon words in each prompt during source fine-tuning might be beneficial. 
For instance, \say{src-5, tgt-0} improves the performance of \say{src-0, tgt-0} by an average of 3.96 F1 points.

\begin{figure}[t]
    \centering
    \includegraphics[width=\columnwidth]{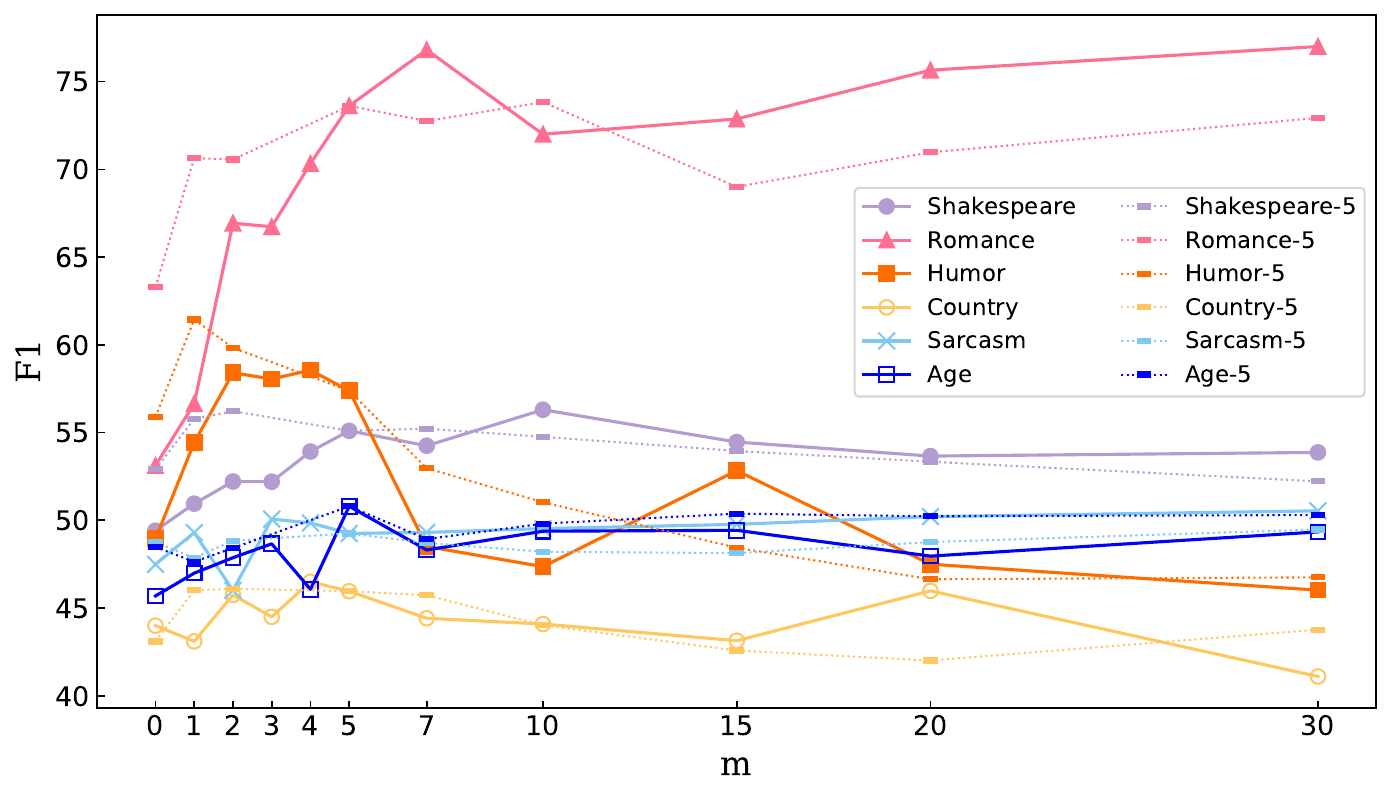}
    \caption{Impact of the number ($m$) of lexicon words or phrases used in each lexicon-based instruction. The solid lines represent the cases where $m$ is applied to both source fine-tuning and target evaluation. The dotted lines (i.e., \emph{Style}-5) show the scores of target styles when lexicon size 5 is used for source fine-tuning, while the size of target-style lexicons $m$ is varied for evaluation.
    }
    \label{appendix:ablation_lex_size_train_test}
\end{figure}

Figure \ref{appendix:ablation_lex_size_train_test} provides a detailed view of the performance change associated with an increase in $m$, broken down by each target style. It reveals that different styles reach their peak performance at different values of $m$. 

\subsection{More Experiments on Style Splits}
\label{appendix:style_splits}
This section presents additional experimental results of our approach, utilizing various style splits outlined in Table \ref{tab:style_splits}. Results are presented in Table \ref{tab:style_splits_results}. It is observed that lexicon-based instruction tuning consistently outperforms standard instruction tuning across various style splits in both T5 and GPT-J models.

\begin{table}[t]
\centering
\small
\begin{tabular}{@{}ll@{}}
\toprule
Split & Source Styles             \\ \midrule
{\styleSplitAa} & \begin{tabular}[c]{@{}l@{}}Politeness, Formality, Sentiment\end{tabular}   \\\midrule
{\styleSplitA} & \begin{tabular}[c]{@{}l@{}}Politeness, Formality, Sentiment,\\ Hate/Offense\end{tabular}                                         \\\midrule
{\styleSplitB}  & \begin{tabular}[c]{@{}l@{}}Politeness, Formality, Sentiment, \\ Hate/Offense, Politics\end{tabular}                               \\\midrule
{\styleSplitC}  & \begin{tabular}[c]{@{}l@{}}Politeness, Formality, Sentiment, \\ Hate/Offense, Politics, Readability, \\ Subjectivity\end{tabular} \\ \bottomrule
\end{tabular}
\caption{Source styles used in different source-target style splits. 
}
\label{tab:style_splits}
\end{table}
\begin{table}[t]
\begin{adjustbox}{width=\linewidth}
\centering
\small
\begin{tabular}{@{}lllllll@{}}
\toprule
Model      & \#Params       & Instruction   & {\styleSplitAa}     & {\styleSplitA}       & {\styleSplitB}       & {\styleSplitC}         \\ \midrule
      &            & {\noLexMethod}    & 36.72  & 36.27                         & 30.01                         & 33.72                         \\
\multirow{-2}{*}{T5}    & \multirow{-2}{*}{220M} & \cellcolor[HTML]{EFEFEF}{\LexPTmethod} &\cellcolor[HTML]{EFEFEF}53.30 & \cellcolor[HTML]{EFEFEF}53.27 & \cellcolor[HTML]{EFEFEF}54.18 & \cellcolor[HTML]{EFEFEF}57.30 \\ \midrule
     &             & {\noLexMethod}  & 50.14    & 53.64                         & 56.06                         & 51.96                         \\
\multirow{-2}{*}{GPT-J} & \multirow{-2}{*}{6B}   & \cellcolor[HTML]{EFEFEF}{\LexPTmethod}&\cellcolor[HTML]{EFEFEF}54.14 & \cellcolor[HTML]{EFEFEF}56.15 & \cellcolor[HTML]{EFEFEF}57.52 & \cellcolor[HTML]{EFEFEF}56.32 \\ \bottomrule
\end{tabular}
\end{adjustbox}
\caption{Average F1 on the six evaluation styles. Across all training-evaluation splits, {\LexPTmethod} instruction improves the average performance on unseen styles compared to {\noLexMethod} instruction for both T5 and GPT-J.}
\label{tab:style_splits_results}
\end{table}

\begin{table*}[t]
\centering
\begin{adjustbox}{width=2\columnwidth}
\begin{tabular}{@{}c|l|l@{}}
\toprule
\textbf{\begin{tabular}[c]{@{}c@{}}Instruction \\ Template\end{tabular}} & \multicolumn{1}{c|}{\textbf{Input}}                                                                                                                                                                                                                                                                                                                                                   & \multicolumn{1}{c}{\textbf{Output}} \\ \midrule
\multirow{5}{*}{t1}                                                   & \multirow{5}{*}{\begin{tabular}[c]{@{}l@{}}Which style best describes the sentence “\{sentence\}”?\\ styles:\\ - \{className$_1$\}: \{$e_1,\cdots, e_k$\}\\ - \{className$_2$\}: \{$e_1,\cdots, e_k$\}\\ ...\end{tabular}}                                                                                                                                                                              & \multirow{13}{*}{className$_i$}       \\
                                                                      &                                                                                                                                                                                                                                                                                                                                                                                       &                                     \\
                                                                      &                                                                                                                                                                                                                                                                                                                                                                                       &                                     \\
                                                                      &                                                                                                                                                                                                                                                                                                                                                                                       &                                     \\
                                                                      &                                                                                                                                                                                                                                                                                                                                                                                       &                                     \\ \cmidrule(r){1-2}
\multirow{3}{*}{t2}                                                   & \multirow{3}{*}{\begin{tabular}[c]{@{}l@{}}In this task, you are given sentences. The task is to classify a sentence as "\{className$_1$\}" if the\\ style of the sentence is similar to the words "\{$e_1,\cdots, e_k$\}" or as "\{className$_2$\}" if the style of\\ the sentence is similar to the words "\{$e_1,\cdots, e_k$\}" or as $\cdots$ Here is the sentence: "\{sentence\}".\end{tabular}} &                                     \\
                                                                      &                                                                                                                                                                                                                                                                                                                                                                                       &                                     \\
                                                                      &                                                                                                                                                                                                                                                                                                                                                                                       &                                     \\ \cmidrule(r){1-2}
\multirow{3}{*}{t3}                                                   & \multirow{3}{*}{\begin{tabular}[c]{@{}l@{}}The task is to classify styles of sentences. We define the following styles: "\{className$_1$\}" is\\ defined by "\{$e_1,\cdots, e_k$\}"; "\{className$_2$\}" is defined by "\{$e_1,\cdots, e_k$\}"; $\cdots$ Here is the sentence:\\ "\{sentence\}", which is more like\end{tabular}}                                                                     &                                     \\
                                                                      &                                                                                                                                                                                                                                                                                                                                                                                       &                                     \\
                                                                      &                                                                                                                                                                                                                                                                                                                                                                                       &                                     \\ \cmidrule(r){1-2}
\multirow{2}{*}{t4}                                                   & \multirow{2}{*}{\begin{tabular}[c]{@{}l@{}}Context: "\{className$_1$\}" is defined by "\{$e_1,\cdots, e_k$\}", "\{className$_2$\}" is defined by "\{$e_1,\cdots, e_k$\}"\\ ... Sentence: \{sentence\} Question: which is the correct style of the sentence? Answer:\end{tabular}}                                                                                                                  &                                     \\
                                                                      &                                                                                                                                                                                                                                                                                                                                                                                       &                                     \\ \bottomrule
\end{tabular}
\end{adjustbox}
\caption{Instruction templates.}
\label{tab:prompts_examples}
\end{table*}

\subsection{Comparisons of MetaICL and Lexicon-Based Instructions in Few-Shot Learning}
\label{appendix:ourMethodWithKtuning}
To compare lexicon-based instructions and MetaICL fairly, it is necessary to incorporate supervision from $K$ demonstrations in evaluation style into our approach. We thus introduce a modification to lexicon-based instructions called {\ourMethodWithKtuning}. Specifically, for each evaluation style, we randomly select $K$ examples from its train set and assign a label to each. Next, a model that was previously fine-tuned on the training styles using the 'Lang, Rw' lexicon-based instructions, is further fine-tuned on these $K$ demonstrations. Finally, we evaluate the model on the evaluation style using lexicon-based instructions without demonstrations. 

The results are reported in Table \ref{appendix:few_shot_results}. It is observed that with random labels, {\ourMethodWithKtuning} generally outperforms other methods. These may suggest that lexicons can provide a useful signal for the prediction of unseen styles when the gold labels of examples are absent. 

\begin{table*}[t]
\begin{adjustbox}{width=\linewidth}
\centering
\small
\begin{tabular}{llccccccc}
\toprule
         & \textbf{Method}    & Shakespeare & Romance & Humor & Country & Sarcasm & Age   & \textbf{Avg.} \\ \midrule
\multicolumn{1}{l}{\multirow{5}{*}{\begin{tabular}[c]{@{}l@{}} Examples w/ \\ \emph{random} labels\end{tabular}}} & MetaICL$_4$    & 44.37$_{\pm6.99}$       & 56.21$_{\pm26.64}$   & 37.82$_{\pm5.02}$ & 41.84$_{\pm18.46}$   & 35.55$_{\pm2.94}$   & 40.96$_{\pm11.19}$ & 42.79     \\
\multicolumn{1}{l}{}    &  \cellcolor[HTML]{EFEFEF}MetaICL$_4$+Lex      & \cellcolor[HTML]{EFEFEF}39.80$_{\pm1.47}$       & \cellcolor[HTML]{EFEFEF}64.58$_{\pm18.72}$   & \cellcolor[HTML]{EFEFEF}38.59$_{\pm4.41}$ & \cellcolor[HTML]{EFEFEF}49.72$_{\pm0.44}$  & \cellcolor[HTML]{EFEFEF}43.77$_{\pm6.52}$   & \cellcolor[HTML]{EFEFEF}35.30$_{\pm0.00}$ & \cellcolor[HTML]{EFEFEF}45.29\\
\multicolumn{1}{l}{}    & +Lex +4  & 54.97$_{\pm0.52}$ &\textbf{83.63$_{\pm4.76}$} &\textbf{58.11$_{\pm2.81}$}  & 49.07$_{\pm0.48}$ &\textbf{47.98$_{\pm0.61}$}  & 46.44$_{\pm0.97}$ & \textbf{56.70}    \\
\multicolumn{1}{l}{}    &\cellcolor[HTML]{EFEFEF}MetaICL$_{16}$    &\cellcolor[HTML]{EFEFEF}55.49$_{\pm11.66}$  &\cellcolor[HTML]{EFEFEF}66.91$_{\pm20.48}$  &\cellcolor[HTML]{EFEFEF}36.11$_{\pm4.58}$ &\cellcolor[HTML]{EFEFEF}7.74$_{\pm4.67}$ &\cellcolor[HTML]{EFEFEF}33.33$_{\pm0.00}$   &\cellcolor[HTML]{EFEFEF}31.24$_{\pm0.00}$ &\cellcolor[HTML]{EFEFEF}38.47         \\
\multicolumn{1}{l}{}  & +Lex +16         &\textbf{56.68}$_{\pm2.71}$       &66.87$_{\pm17.72}$ &57.69$_{\pm1.93}$ & \textbf{51.67$_{\pm0.76}$ }   &45.67$_{\pm3.71}$  &\textbf{47.81$_{\pm1.62}$} &54.40        \\
\midrule
\multicolumn{1}{l}{\multirow{5}{*}{\begin{tabular}[c]{@{}l@{}} Examples w/ \\ \emph{gold} labels\end{tabular}}}   &\cellcolor[HTML]{EFEFEF}MetaICL$_4$  & \cellcolor[HTML]{EFEFEF}64.30$_{\pm13.01}$       & \cellcolor[HTML]{EFEFEF}53.53$_{\pm27.30}$   & \cellcolor[HTML]{EFEFEF}49.79$_{\pm12.46}$ & \cellcolor[HTML]{EFEFEF}49.29$_{\pm0.01}$   & \cellcolor[HTML]{EFEFEF}34.28$_{\pm1.57}$   & \cellcolor[HTML]{EFEFEF}36.21$_{\pm1.25}$ & \cellcolor[HTML]{EFEFEF}47.90         \\
\multicolumn{1}{l}{}    &MetaICL$_4$+Lex      &43.90$_{\pm8.06}$       &75.80$_{\pm6.52}$   & 42.78$_{\pm3.99}$ &49.42$_{\pm0.36}$  &38.62$_{\pm3.69}$  & 35.30$_{\pm0.00}$ & 47.63      \\
\multicolumn{1}{l}{}    &\cellcolor[HTML]{EFEFEF} +Lex +4   &\cellcolor[HTML]{EFEFEF}54.42$_{\pm1.78}$    & \cellcolor[HTML]{EFEFEF}85.48$_{\pm3.00}$& \cellcolor[HTML]{EFEFEF}58.83$_{\pm4.93}$ & \cellcolor[HTML]{EFEFEF}48.92$_{\pm0.43}$  & \cellcolor[HTML]{EFEFEF}43.11$_{\pm4.91}$  & \cellcolor[HTML]{EFEFEF}45.84$_{\pm1.94}$ & \cellcolor[HTML]{EFEFEF}56.10    \\
\multicolumn{1}{l}{}    &  MetaICL$_{16}$          & \textbf{72.93}$_{\pm8.15}$       & \textbf{95.79}$_{\pm0.84}$   & 52.05$_{\pm8.52}$& 47.90$_{\pm3.07}$ &  33.33$_{\pm0.00}$  &  35.30$_{\pm0.00}$ & 56.22    \\
\multicolumn{1}{l}{}    &  \cellcolor[HTML]{EFEFEF}+Lex +16 & \cellcolor[HTML]{EFEFEF}60.99$_{\pm6.75}$        & \cellcolor[HTML]{EFEFEF}94.00$_{\pm1.41}$   &\cellcolor[HTML]{EFEFEF}\textbf{63.26}$_{\pm3.35}$  & \cellcolor[HTML]{EFEFEF}\textbf{51.85}$_{\pm0.41}$ & \cellcolor[HTML]{EFEFEF}\textbf{44.93}$_{\pm4.34}$ & \cellcolor[HTML]{EFEFEF}\textbf{47.42}$_{\pm4.54}$  & \cellcolor[HTML]{EFEFEF}\textbf{60.41}   \\ \bottomrule
\end{tabular}
\end{adjustbox}
\caption{Few-shot learning of GPT-J. The subscript of MetaICL represents the number (K) of demonstrations in one prompt.
For each method (MetaICL$_{\text{K}}$, MetaICL$_{\text{K}}$+Lex, or {\ourMethodWithKtuning}), we choose a set of $K$ examples with five different random seeds. By introducing lexicons into prompts, the standard deviation of performance across five runs generally decreases. 
}
\label{appendix:few_shot_results}
\end{table*}

\subsection{Varying Number of Training Examples (K) used in Few-Shot Learning}
\label{appendix:ablation_k}
\begin{figure}[t]
    \centering
    \includegraphics[width=\columnwidth]{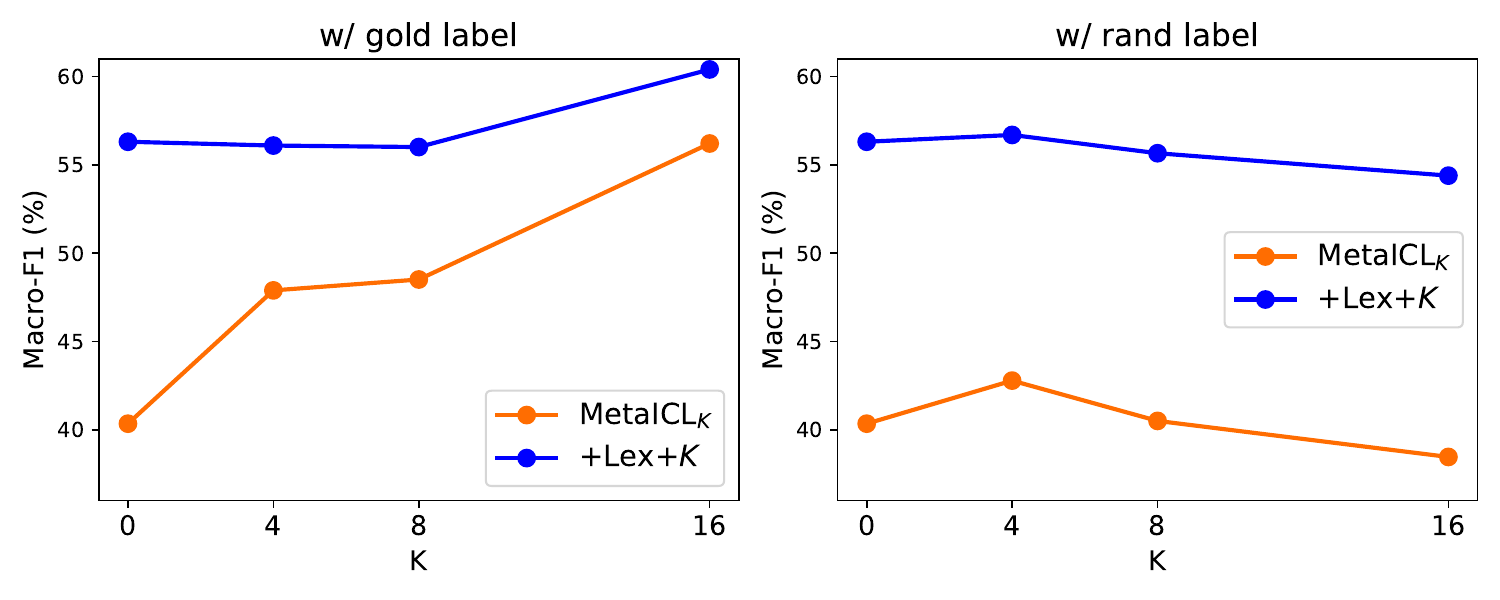}
    \caption{Ablation on the number of training examples ($\text{K}$) in a few-shot learning setting.}
    \label{fig:ablation_k}
\end{figure}

We investigate the impact of the number of examples ($K$) that are used in the few-shot learning methods MetaICL$_\text{K}$ and {\ourMethodWithKtuning}.
Results are reported in Figure \ref{fig:ablation_k}.
The performance of both methods deteriorates with an increase in $K$ when using random labels. However, when gold labels are utilized for the target-style training examples, the performance improves with larger $K$, particularly showing significant improvement from $K=8$ to $K=16$. Moreover, as $K$ increases, the performance disparity between utilizing ground-truth labels and random labels further expands. These observations show that the ground-truth input-label mapping is important in our case.

\section{More Prompting Examples}
Figure \ref{fig:lp_examples} shows the example input and output for all lexicon-based instruction variants. In MetaICL$_\text{K}$+Lex, one prompt consists of $K$ demonstrations and an input sentence. Figure \ref{fig:metaicl+lex_example.png} provides an example demonstration.

\end{document}